%% file: main.tex
\documentclass[11pt, a4paper, logo, copyright, nonumbering]{antgroup}
\usepackage[square, numbers]{natbib}
\usepackage{dblfloatfix}
\usepackage{ulem}
\usepackage{caption}
\usepackage{dramatist}
\usepackage{xspace}
\usepackage{pifont}
\usepackage{multirow}

\usepackage{xltabular}
\usepackage{longtable}
\usepackage{hyperref}
\usepackage{amsfonts}
\usepackage{amsmath}
\usepackage{amssymb}
\usepackage{lineno}
\usepackage{multirow}
\usepackage{adjustbox}

\usepackage[bottom]{footmisc}

\usepackage{CJKutf8}
\usepackage{subfigure}
\usepackage{setspace}

\usepackage{dsfont}
\usepackage{array}
\usepackage{tabularx}
\usepackage{subfigure}
\usepackage{xcolor}
\usepackage{tabularx}
\usepackage{booktabs}

\usepackage{lipsum}
\usepackage{multicol}
\usepackage{wrapfig}
\usepackage[most,skins,theorems]{tcolorbox}
\usepackage{array}
\definecolor{customgray}{RGB}{230,230,230}
\usepackage{subfiles}

\usepackage{tabularray}
\usepackage{xcolor}
\usepackage{amssymb}
\usepackage{xcolor}
\usepackage{wrapfig}
\usepackage{afterpage}
\usepackage{enumitem}
\usepackage{microtype}

\newcommand{\fullstar}{\textcolor{orange!80!black}{\ding{72}}}
\newcommand{\emptystar}{\textcolor{gray!40}{\ding{73}}}

\newcommand{\stars}[1]{%
  \ifnum#1>0\fullstar\else\emptystar\fi
  \ifnum#1>1\fullstar\else\emptystar\fi
  \ifnum#1>2\fullstar\else\emptystar\fi
  \ifnum#1>3\fullstar\else\emptystar\fi
  \ifnum#1>4\fullstar\else\emptystar\fi
}

\tcbset{
  aibox/.style={
    width=\linewidth,
    top=8pt,
    bottom=4pt,
    colback=gray!10!white,
    colframe=gray!50!black,
    colbacktitle=gray!70!black,
    enhanced,
    center,
    attach boxed title to top left={yshift=-0.1in,xshift=0.15in},
    boxed title style={boxrule=0pt,colframe=white,},
  }
}
\newtcolorbox{AIbox}[2][]{aibox,title=#2,#1}

\makeatletter
\def\@BTrule[#1]{%
  \ifx\longtable\undefined
    \let\@BTswitch\@BTnormal
  \else\ifx\hline\LT@hline
    \nobreak
    \let\@BTswitch\@BLTrule
  \else
     \let\@BTswitch\@BTnormal
  \fi\fi
  \global\@thisrulewidth=#1\relax
  \ifnum\@thisruleclass=\tw@\vskip\@aboverulesep\else
  \ifnum\@lastruleclass=\z@\vskip\@aboverulesep\else
  \ifnum\@lastruleclass=\@ne\vskip\doublerulesep\fi\fi\fi
  \@BTswitch}
\makeatother

\addto\extrasenglish{
}

 {\begin{list}{}%
         {\setlength{\leftmargin}{#1}}%
         \item[]%
 }
 {\end{list}}

\reportnumber{001} %

\renewcommand{\today}{}

\title{\centering AoE: Always-on Egocentric Human Video Collection for Embodied AI}

\renewcommand{\thefootnote}{\fnsymbol{footnote}}

\author{%
    \begin{minipage}{\textwidth}
    \centering
    Bowen Yang$^{*1}$,
    Zishuo Li$^{*1}$,
    Yang Sun$^{*1}$,
    Changtao Miao$^{*\dagger 1}$,
    Yifan Yang$^{2}$,
    Man Luo$^{1}$,
    Xiaotong Yan$^{1}$,
    Feng Jiang$^{1}$,
    Jinchuan Shi$^{3}$,
    Yankai Fu$^{4}$,
    Ning Chen$^{4}$,
    Junkai Zhao$^{5}$,
    Pengwei Wang$^{5}$,
    Guocai Yao$^{5}$,
    Shanghang Zhang$^{4,5}$,
    Hao Chen$^{3}$,
    Zhe Li$^{1}$,
    Kai Zhu$^{\ddagger 1}$

    \smallskip
    {\small
    $^{1}$Ant Digital Technologies, Ant Group \quad
    $^{2}$Institute of Automation, Chinese Academy of Sciences

    \smallskip
    {\small
    $^{3}$Zhejiang University} \quad
    $^{4}$Peking University \quad
    $^{5}$Beijing Academy of Artificial Intelligence}

    \smallskip
    {\small $^{\ddagger}$\href{kaile.zk@antgroup.com}{kaile.zk@antgroup.com}}
    \end{minipage}%
}

\input{commands.tex}

\begin{abstract}

Embodied foundation models require large-scale, high-quality real-world interaction data for pre-training and scaling.
However, existing data collection methods suffer from high infrastructure costs, complex hardware dependencies, and limited interaction scope, making scalable expansion challenging.
In fact, humans themselves are ideal physically embodied agents. Therefore, obtaining egocentric real-world interaction data from globally distributed “human agents” offers advantages of low cost and sustainability.
To this end, we propose the \textbf{\textit{Always-on Egocentric (AoE)}} data collection system, which aims to simplify hardware dependencies by leveraging humans themselves and their smartphones, enabling low-cost, highly efficient, and scene-agnostic real-world interaction data collection to address the challenge of data scarcity.
Specifically, we first employ an ergonomic neck-mounted smartphone holder to enable low-barrier, large-scale egocentric data collection through a cloud-edge collaborative architecture. Second, we develop a cross-platform mobile \textit{APP} that leverages on-device compute for real-time processing, while the cloud hosts automated labeling and filtering pipelines that transform raw videos into high-quality training data. Finally, the \textbf{\textit{AoE}} system supports distributed Ego video data collection by \textbf{\textit{anyone}}, \textbf{\textit{anytime}}, and \textbf{\textit{anywhere}}.
We evaluate \textbf{\textit{AoE}} on data preprocessing quality and downstream tasks, demonstrating that high-quality egocentric data significantly boosts real-world generalization.

\end{abstract}

\begin{document}
\maketitle

\begingroup
  \renewcommand{\thefootnote}{}
  \footnotetext{%
    $^*$Equal Contribution.\quad
    $^\dagger$Corresponding Author.\quad
    $^\ddagger$Project Leader.%
  }
\endgroup

\begin{figure*}[htbp]
\centering
\includegraphics[width=0.99\textwidth]{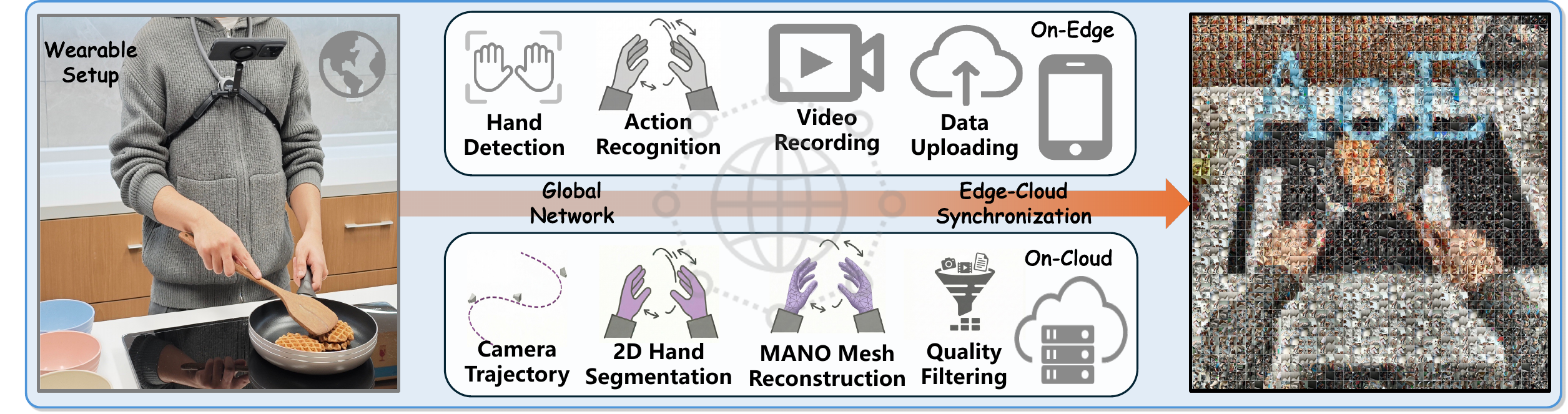}
\caption{
\textbf{Overview of the \textit{AoE} system.}
The system leverages neck-mounted smartphones for ubiquitous egocentric capture (\textit{Left}). Our edge-cloud collaborative pipeline (\textit{Middle}) efficiently distributes computation: on-device models handle real-time detection and selective uploading, while cloud servers execute heavy-duty auto-labeling and quality filtering. This design minimizes hardware dependencies, enabling scalable, high-quality data collection in the wild (\textit{Right}).
}
\label{figure:fig1}
\end{figure*}


\subfile{Sections/1_introduction.tex}

\subfile{Sections/2_related_work.tex}

\subfile{Sections/3_method.tex}

\subfile{Sections/4_experiments.tex}

\subfile{Sections/5_conclusion.tex}

\clearpage
\bibliographystyle{unsrtnat} 
\bibliography{main}

\newpage

\subfile{Sections/6_Appendix.tex}

\end{document}

%% file: commands.tex
\renewcommand{\phi}{\varphi}

\renewcommand{\epsilon}{\varepsilon}
\renewcommand{\imath}{\mathrm{i}}

\newlength{\restsubwidth}
\newlength{\restsubheight}
\newlength{\restsubmoreheight}
\newcommand{\rest}[2]{%
        \settowidth{\restsubwidth}{\ensuremath{#2}}
        \settoheight{\restsubheight}{\ensuremath{{}_{#2}}}
        \ensuremath{{#1\hskip 0.5pt}_{\vrule\kern2pt\parbox[b][%
        4pt][b]{\the\restsubwidth}{%
                        \ensuremath{{}_{#2}}}}}
        }

%% file: Sections/1_introduction.tex
\section{Introduction}

Recent strides in embodied foundation models, exemplified by Gen-0 ~\citep{generalist2025gen0}, exhibit ``scaling laws'' mirroring those observed in Large Language Models (LLMs): model generalization capabilities and performance frontiers expand commensurate with the scale of real-world interaction data ~\citep{generalist2025gen0,luo2025being}. This trend underscores that data-driven paradigms have become the cornerstone for advancing embodied intelligence. However, in stark contrast to the ubiquitous textual corpora available on the web, physical interaction data is notoriously difficult to curate, suffering from prohibitive acquisition costs, limited diversity, and significant inherent noise. Consequently, the scarcity of large-scale, high-quality real-world interaction data remains a critical bottleneck impeding the field of embodied intelligence.

\begin{table}[tp]
\caption{
\textbf{Comparison of data collection paradigms for dexterous manipulation.} The \textbf{\textit{AoE}} retains the high scalability and low cost ($<\$20$) compares with Teleoperation, UMIs~\citep{chi2024universal,xu2025dexumi}, Wearables~\citep{zhong2025humanoidexo}, Passive Videos~\citep{grauman2022ego4d}. 
Detailed rating criteria are provided in the Appendix~\ref{appendix:rating_criteria}.
}
\label{tab:collection_comparison}
\begin{center}
\resizebox{\textwidth}{!}{
    \begin{tabular}{lccccc}
    \toprule
     & \textbf{Teleoperation} & \textbf{UMIs} & \textbf{Wearables} & \textbf{Passive Videos} & \textbf{\textit{AoE} (Ours)} \\
    \midrule
    Cost (per user)      & $>$\$50k & \$300–800 & $>$\$2k & Free & $<$\$20 \\
    \midrule
    Non-Intrusiveness    & \stars{1} & \stars{3} & \stars{1} & \stars{5} & \stars{4} \\
    Scalability          & \stars{1} & \stars{3} & \stars{2} & \stars{5} & \stars{5} \\
    Deployment Ease      & \stars{1} & \stars{2} & \stars{2} & \stars{5} & \stars{5} \\
    Data Quality         & \stars{5} & \stars{4} & \stars{4} & \stars{1} & \stars{4} \\
    \bottomrule
    \end{tabular}
}
\vspace{-10pt}
\end{center}
\end{table}

Early teleoperation-based approaches ~\citep{chen2025towards,luo2025human} capture high-fidelity data but are constrained by expensive hardware and lab environments, impeding large-scale expansion. To enable in-the-wild collection, researchers have developed portable solutions, including handheld interfaces ~\citep{chi2024universal,zhaxizhuoma2025fastumi,xu2025dexumi} and wearable AR/VR systems ~\citep{wang2024dexcap,zhong2025humanoidexo,fang2024airexo,zeng2025activeumi}. However, handheld devices require active manipulation that disrupts natural behavior, while bulky AR/VR setups remain intrusive, rendering them unsuitable for continuous daily use.
Alternatively, learning directly from egocentric videos ~\citep{luo2025being,yang2025egovla} offers a scalable avenue for pre-training Vision-Language-Action (VLA) models. Nevertheless, existing datasets ~\citep{grauman2022ego4d,damen2018scaling} typically lack the fine-grained interaction dynamics essential for policy learning. Consequently, extracting high-quality segments from such noisy streams incurs substantial curation costs, limiting their practical utility for robot training, as shown in Table \ref{tab:collection_comparison}.

Humans naturally function as ideal physically embodied agents, offering a massive population continuously operating within the environments we seek to capture~\citep{yang2025egovla,qiu2025humanpolicy}. This mirrors data acquisition paradigms in autonomous driving, where fleets aggregate environmental signals and behavioral data globally~\citep{ramanishka2018CVPR,mao2021one}. Inspired by this, we propose harvesting egocentric interaction data from these distributed "human agents" in unstructured, everyday scenarios. Crucially, ubiquitous smartphones provide a low-cost, deployable foundation for visual sensing and edge computing. Consequently, organically coupling human dexterity with mobile computing to establish a human-centric, distributed collection architecture constitutes a scalable and sustainable technical pathway.

To this end, we introduce the \textbf{\textit{Always-on Egocentric (AoE) system}}, a framework leveraging ubiquitous smartphones to enable continuous, low-cost, and scalable egocentric video acquisition, as illustrated in Figure \ref{figure:fig1}. Adhering to a human-centered design, \textbf{\textit{AoE}} utilizes an ergonomic, magnetic neck mount that stabilizes the device at the chest. This setup allows the rear camera to capture physical interactions in natural environments while minimizing disruption to daily activities, thereby preserving behavioral naturalness and data fidelity.
Structurally, the framework constitutes a loosely coupled distributed architecture linking global cloud servers with edge devices, significantly lowering technical barriers for large-scale contribution. We develop a cross-platform mobile \textit{APP} (Android/iOS) that exploits smartphone capabilities for real-time calibration and pre-processing, eliminating reliance on specialized hardware. Finally, we establish a robust cloud-based pipeline for automated annotation and filtering. Upon user-authorized upload, this edge-cloud system performs computation-intensive tasks to transform raw, noisy video streams into high-quality training segments characterized by high information density and clear semantics.

We conduct a comprehensive evaluation of the \textbf{\textit{AoE}} data acquisition system across three key dimensions: data preprocessing efficacy, real-to-sim reconstruction fidelity, and downstream policy fine-tuning. Experimental results demonstrate that our collected egocentric interaction data exhibit high-quality annotations and robust reconstruction capabilities. Crucially, leveraging this data significantly enhances the task success rates of embodied models in complex real-world environments.

To summarize, our key contributions are the following:
\begin{itemize}
\item We propose the \textbf{\textit{Always-on Egocentric (AoE)}} system, a scalable, low-cost framework utilizing ubiquitous smartphones and ergonomic hardware to enable continuous, non-intrusive data acquisition in the wild.
\item We design a \textbf{distributed edge-cloud architecture} featuring an automated pipeline that efficiently transforms raw, noisy video streams into high-density, semantically annotated training segments for embodied learning.
\item We demonstrate that AoE-collected data significantly enhances \textbf{real-to-sim reconstruction fidelity} and improves the task success rates of downstream \textbf{robotic manipulation policies}.
\end{itemize}

%% file: Sections/2_related_work.tex
\section{Related Works}

\subsection{Data Collection with In-The-Wild Equipment}

Large-scale "in-the-wild" data is vital for embodied intelligence to transcend hardware and lab constraints~\citep{lindata,generalist2025gen0,chen2025towards}. We categorize existing methods by acquisition paradigm:
\textbf{(1) Handheld Gripper-based Interfaces:} While UMI~\citep{chi2024universal} and successors (e.g., FastUMI~\citep{zhaxizhuoma2025fastumi}, DexUMI~\citep{xu2025dexumi}) lower costs via portable grippers, they remain "active" devices requiring deliberate hardware manipulation, thereby restricting natural interaction coverage.
\textbf{(2) Wearable AR/VR Systems:} High-precision setups like DexCap~\citep{wang2024dexcap}, HumanoidExo~\citep{zhong2025humanoidexo}, AirExo~\citep{fang2024airexo}, and VR-based hubs (e.g., exUMI~\citep{xu2025exumi}, ActiveUMI~\citep{zeng2025activeumi}) capture complex poses. However, their significant bulk ($>500g$) and power dependencies render them too intrusive for continuous, "always-on" daily use.
\textbf{(3) Passive Egocentric Videos:} Existing datasets (e.g., Ego4D~\citep{grauman2022ego4d} and Epic-Kitchens~\citep{damen2018scaling}) prioritize recognition over manipulation, suffering from motion blur and occlusion. Consequently, extracting training-grade segments incurs prohibitive curation costs.
Therefore, we introduce \textbf{\textit{AoE}} (Always-On Egocentric), a system leveraging ubiquitous smartphones for passive, scalable, and high-quality data acquisition with minimal user burden and near-zero marginal cost.

\subsection{Manipulation Policies Learning from Human Videos}

Leveraging human videos addresses robot data scarcity. Early efforts focused on representation pretraining~\citep{mavip,nair2023r3m,radosavovic2023real} or extracting proxy signals like trajectories~\citep{bharadhwaj2024track2act,wen2023any} and plans~\citep{wangmimicplay}. Recent policy learning approaches span three paradigms: (1) bridging embodiment gaps via retargeting on controlled data~\citep{qin2022dexmv,chen2022dextransfer,chen2025vividex}; (2) utilizing latent action codes to bypass explicit labeling~\citep{yelatent,bu2025univla,bjorck2025gr00t}; and (3) recovering 3D hand motion to construct shared action spaces~\citep{mandikal2022dexvip,yang2025egovla,luo2025being,bi2025h}. Crucially, while recent studies~\citep{cai2025n,li2025vitra} validate in-the-wild pretraining, they underscore the bottleneck of curating noisy corpora. Our \textbf{\textit{AoE}} system resolves this scaling challenge by enabling low-cost, automated acquisition and high-quality filtering for robust policy learning.

%% file: Sections/3_method.tex
\section{Always-On Egocentric Data Collection System}

\begin{figure*}[t]
\vspace{-15pt}
\centering
\includegraphics[width=1.0\textwidth]{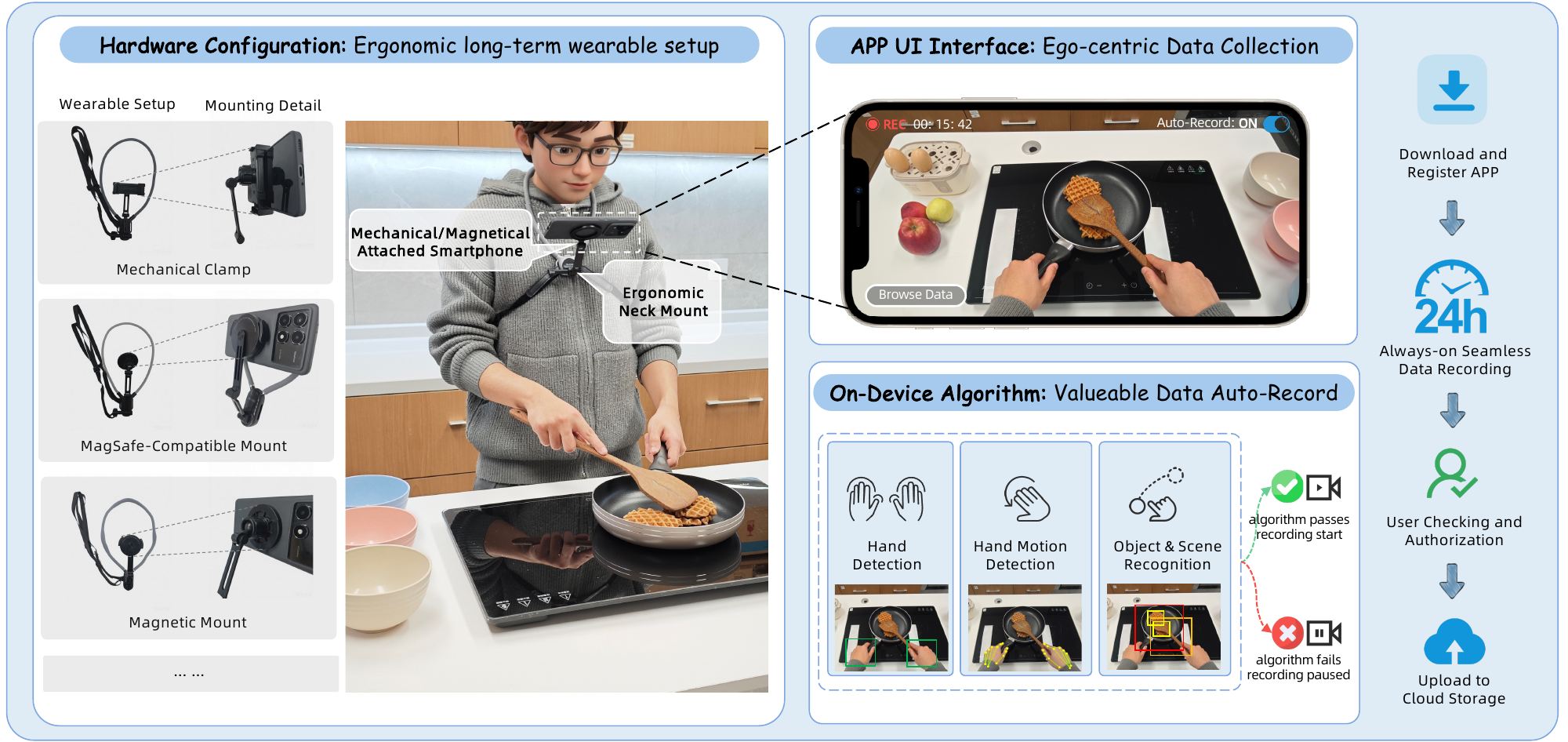}
\caption{
\textbf{Overview of Hardware \& Mobile Application.}
The AOE hardware supports various ergonomic mounts (Mechanical, MagSafe, Magnetic) with stabilizing straps for robust, all-day egocentric recording \textit{(Left)}.
The user-friendly UI Interface for users to manage recordings \textit{(Up Right)}. On-device intelligence selectively records high-value manipulation data \textit{(Bottom Right)}. 
Secure pipeline that synchronizes user-authorized data to the cloud \textit{(Right)}.
}
\vspace{-15pt}
\label{figure:hardware-app}
\end{figure*}






The proposed \textbf{\textit{Always-on Egocentric (AOE)}} system is a distributed edge-cloud framework designed for the scalable, low-cost acquisition of egocentric robot learning data. \textbf{\textit{AoE}} seamlessly captures in-the-wild activities without disrupting user workflows, automatically filtering and annotating high-quality samples. The system prioritizes three objectives: \textbf{(1) Effortless Collection: } Leveraging ubiquitous smartphones and a minimalistic neck mount to democratize data acquisition across diverse environments.
\textbf{(2) High-Quality Annotation:} Employing an automated pipeline to yield training-ready data with negligible manual intervention.
\textbf{(3) Scalable Architecture:} Utilizing edge-cloud collaboration to support high-concurrency streaming and distributed data production.
Subsequent sections detail our hardware design, mobile software, annotation pipeline, and system implementation.

\subsection{Hardware \& Mobile Application}

\textbf{Hardware Configuration and Accessibility.}
While UMI~\citep{chi2024universal} and exoskeletons~\citep{zhong2025humanoidexo} successfully decouple collection from specific embodiments, their reliance on specialized, costly hardware hinders "in-the-wild" scalability. We instead propose a democratized ecosystem built entirely on consumer-grade devices.
As shown in Figure~\ref{figure:hardware-app}, our ergonomic neck mounts support \textbf{mechanical}, \textbf{MagSafe}, and \textbf{magnetic} interfaces. An auxiliary \textbf{stabilizing strap} minimizes camera shake during dynamic manipulation. By mounting smartphones at the sternum and leveraging \textbf{ultra-wide lenses}, we capture high-resolution egocentric views that approximate human vision without obstructing natural workflows or comfort.
Table~\ref{tab:collection_comparison} highlights our decisive cost efficiency: unlike UMI or wearables (\$300--\$2000), our assembly costs \textbf{under \$20}. This near-zero marginal cost, combined with smartphone ubiquity, facilitates highly scalable data acquisition across diverse real-world scenarios.

\textbf{Always-On Collection App.}
We developed a cross-platform application (Android/iOS) featuring an authorized "Always-On" mode for efficient, privacy-preserving data capture. 
As shown in Figure~\ref{figure:hardware-app}), we leverage lightweight on-device vision models, which includes hand detection, motion tracking, and open-set recognition, the system autonomously triggers recording only during relevant hand-object interactions. 
This selective strategy eliminates manual intervention, enabling seamless "24h" collection while minimizing storage and downstream filtering overhead \footnote{\textbf{This mode is only enabled upon explicit user authorization, ensuring the user maintains full control over the collection periods.}}. Additionally, the app captures camera intrinsics and sensor metadata to facilitate automated cloud labeling. The user interface and workflow are detailed in Figure~\ref{figure:app-workflow} and Appendix~\ref{appendix:app_workflow}.

\textbf{Privacy and Data Security.}
Adopting a privacy-first architecture, our system performs all model inference and raw data storage locally. To safeguard subjects, an automated pipeline de-identifies sensitive attributes (e.g., faces, text) prior to upload, while a continuous, non-mutable audio signal alerts bystanders during recording. Finally, complete data sovereignty is ensured by requiring manual user review and authorization before any cloud transmission.


\begin{figure}[t]
\centering
\includegraphics[width=0.99\textwidth]{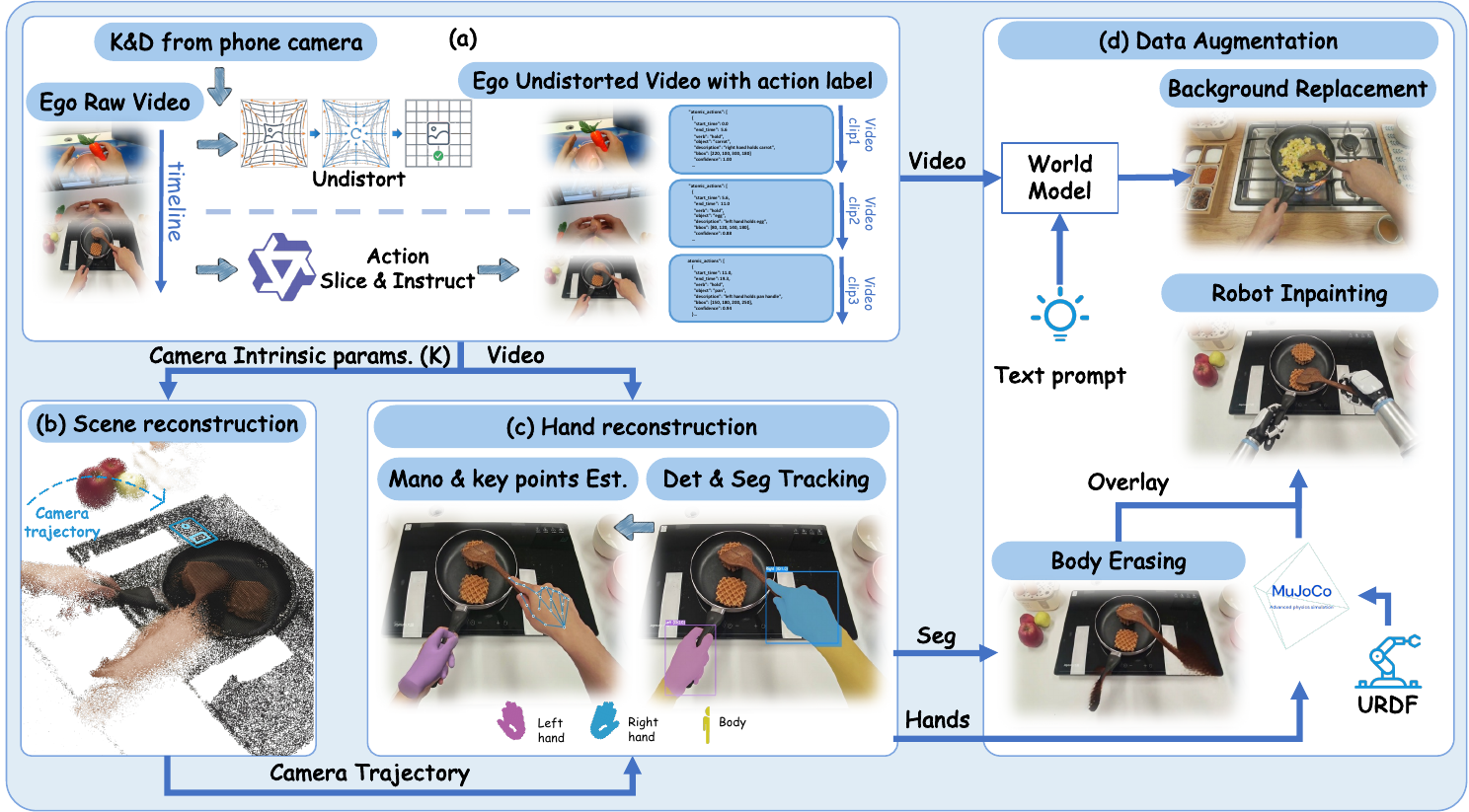}
\caption{
\textbf{Overview of the Automatic Annotation and Augmentation Pipeline.} (a) Undistort videos and segment videos into atomic clips. (b) Dense depth maps yield camera trajectories and scene reconstruction. (c) Hand poses are generated and transformed to world coordinates. (d) Augmentation employs generative background replacement and simulation-based robot inpainting.
}
\vspace{-15pt}
\label{figure:annotation-pipline}
\end{figure}

\subsection{Automated Annotation and Quality Filtering Pipeline}
To address perceptual aliasing and limited VLA compliance in long-horizon tasks~\citep{kim2025fine}, AoE generates cost-efficient atomic-action annotations—comprising end-effector trajectories, camera trajectories and semantic labels via a 6-stage pipeline:

\textbf{Camera Calibration}: We retrieve factory-calibrated intrinsics via Camera2~\citep{android2024camera2overview}, eliminating device-specific calibration overhead while maintaining sub-pixel stability across conditions.

\textbf{Atomic Action Segmentation}: Qwen3-VL-235B-A22B~\citep{Qwen2.5-VL} segments videos into semantic atomic clips (Figure\ref{figure:annotation-pipline}a) avoiding heuristic biases. Human verification corrects potential VLM hallucinations in boundaries and labels.

\textbf{Camera Trajectory Estimation}: We use Lingbot-Depth~\citep{lingbot-depth2026} to refine RGBD inputs with superior stability, while temporal consistency enhances RGB-only metric precision~\citep{wang2025moge2}. Trajectories are estimated via MegaSAM~\citep{li2024megasam} using robust kernels tuned for motion blur (Figure~\ref{figure:annotation-pipline}b).

\textbf{Hand Reconstruction}: HaWoR~\citep{zhang2025hawor} performs 3D joint recovery, rescaled via depth estimates. MANO~\citep{Romero_2017} outputs are transformed to world coordinates via SLAM poses, employing sliding-window optimization to enforce kinematic consistency (Figure~\ref{figure:annotation-pipline} c).

\textbf{Data Augmentation}: Hand replacement follows Masquerade~\citep{lepert2025masquerade}: GAN-based removal, 6-DoF robot alignment, and photometric re-rendering. Background replacement employs video diffusion on contact-free frames conditioned on depth and segmentation (Figure~\ref{figure:annotation-pipline} d).

\textbf{Quality Control}: We automatically filter kinematic outliers ($>3\sigma$ joint velocities) and high reprojection errors ($>5$px). A 5\% manual inspection informs adaptive pipeline adjustments (e.g., texture diversity, hand proximity), with failed samples categorized into a hard-negative pool for re-annotation, effectively closing the data generation loop.

\begin{figure}[t]
\vspace{-15pt}
\centering
\includegraphics[width=0.99\textwidth]{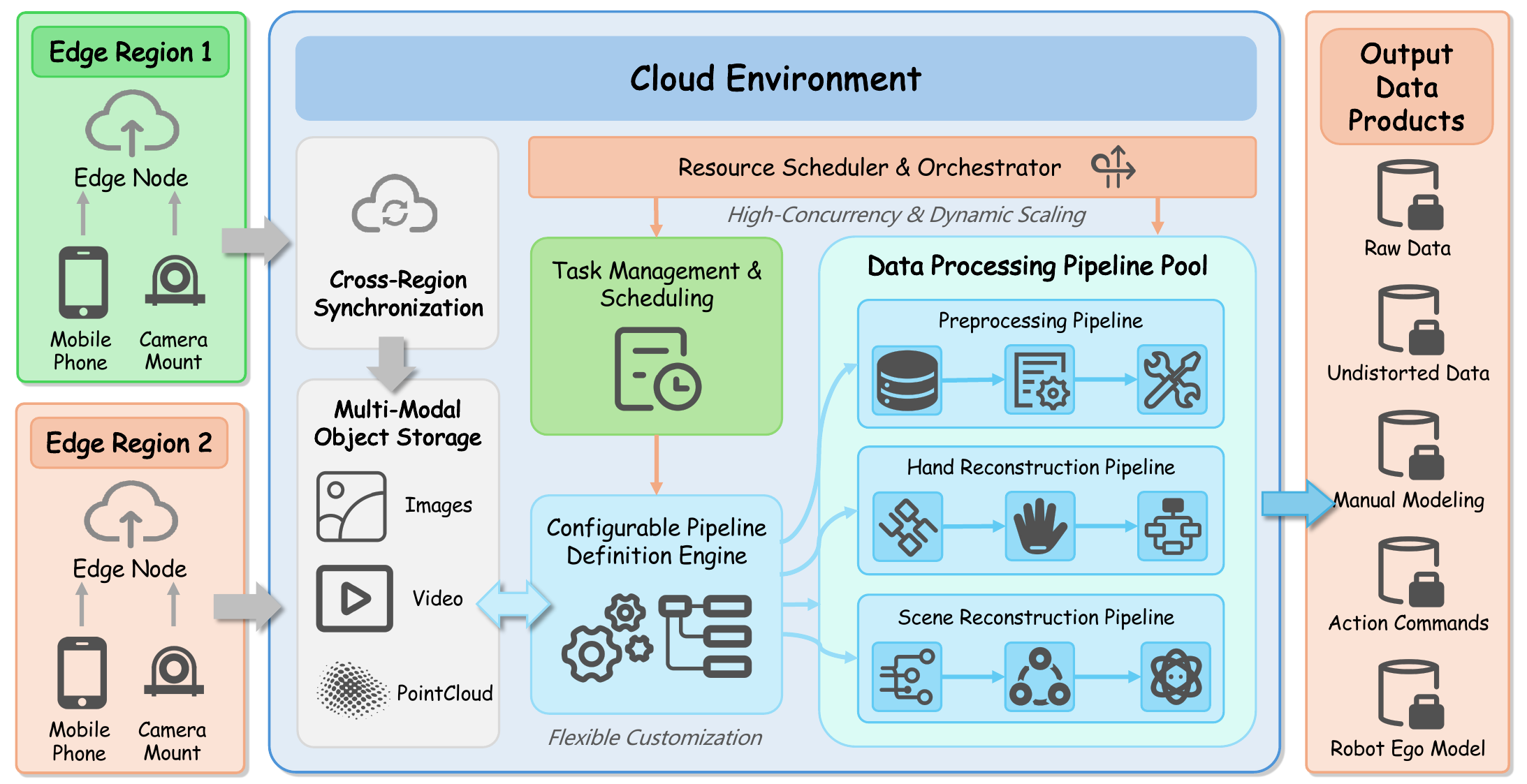}
\caption{
\textbf{Distributed Edge-Cloud Architecture.} Enabling low-latency edge-to-cloud synchronization, the system utilizes a configurable, elastically scaled pipeline to generate multi-modal data for robot policy learning.
}
\vspace{-15pt}
\label{figure:server-pipline}
\end{figure}

\subsection{Distributed System Implementation}

To facilitate large-scale robot learning from human demonstrations across geographically distributed devices, we propose a distributed edge-cloud collaborative architecture. This system is explicitly designed to overcome three critical bottlenecks inherent in traditional centralized data collection: high-latency data transfer, inflexible processing pipelines, and limited scalability. As illustrated in Figure~\ref{figure:server-pipline}, our unified platform establishes a scalable infrastructure for the full lifecycle of egocentric manipulation data by addressing these challenges through three specific architectural innovations.

First, to mitigate high-latency transfer from dispersed collectors, we implement an \textbf{Edge-Cloud Collaborative Architecture}. This component deploys proximity-based edge ingestion nodes equipped with intelligent routing to minimize transfer latency, while simultaneously maintaining centralized data consistency through asynchronous cross-region synchronization. Second, addressing the rigidity of pipelines that typically require weeks of re-engineering for new algorithms, we develop a \textbf{Customizable Processing Pipeline}. By leveraging declarative workflow orchestration, this design enables modular algorithm integration and hot-swappable components, allowing for rapid updates without system-level modifications. Third, to resolve the inefficiencies of fixed resource allocation, we introduce \textbf{Elastic Scaling} mechanisms. Built on a cloud-native Kubernetes architecture, this module utilizes horizontal pod autoscaling and intelligent GPU/CPU resource partitioning to dynamically provision compute resources in response to real-time demand.

Collectively, these innovations yield significant performance improvements: data transfer latency is reduced from 500ms+ to under 100ms; algorithm integration cycles are shortened from weeks to days; and the system supports thousands of concurrent devices with minute-level responses to workload spikes. This architecture effectively enables ``anyone, anytime, anywhere'' data contribution, thereby advancing the scalability of robot manipulation policy learning.
The more details in Appendix~\ref{appendix:distributed_system}.

%% file: Sections/4_experiments.tex
\section{Experiments}

We empirically evaluate the \textbf{\textit{AoE}} system across three core dimensions: (1) validating the precision of 3D hand pose and camera trajectory tracking; (2) assessing the fidelity of interaction reconstruction for real-to-sim transfer; and (3) quantifying downstream utility via measurable performance gains in real-world robotic manipulation tasks.
The more experimental setups in Appendix~\ref{appendix:experoments}.

\begin{figure*}[t]
\vspace{-10pt}
\centering
\includegraphics[width=1.0\textwidth]{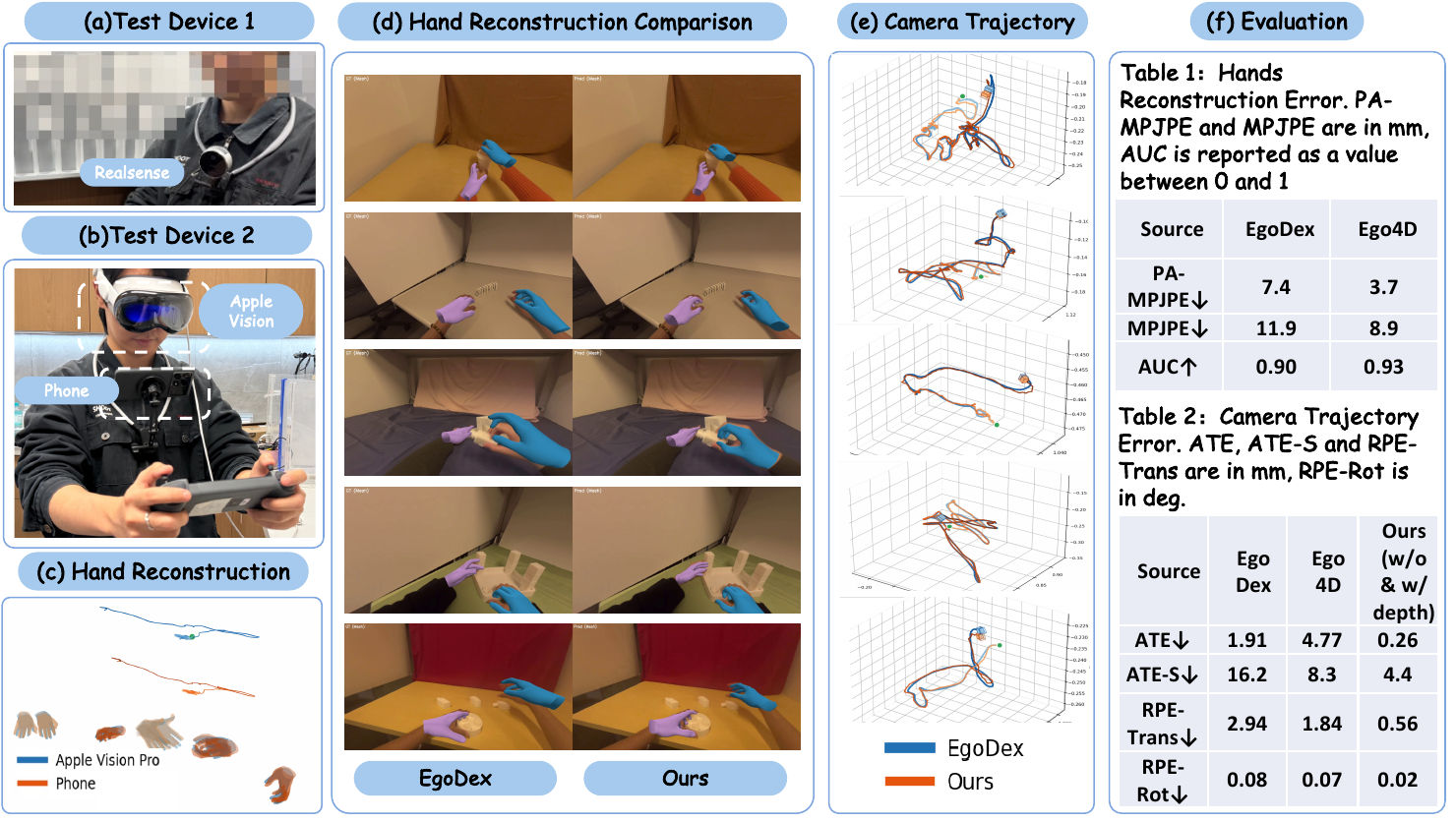}
\caption{
\textbf{Comparison of data processing acquisition methods and accuracy.}
(a) Depth camera acquisition configuration. (b) AR glasses + smartphone combined acquisition configuration. (c) Hand modeling comparison using AR glasses + smartphone. (d) Hand modeling comparison from EgoDex. (e) Camera trajectory comparison after rotation-translation alignment from EgoDex. (f) Hand reconstruction accuracy and camera trajectory reconstruction accuracy.
}
\vspace{-17pt}
\label{fig:pipeline_evaluation}
\end{figure*}

\subsection{Precision of the \textbf{\textit{AoE}} System}
\paragraph{Experimental Setup.} 
We evaluate EgoFactory on four datasets: 
(1) \textbf{EgoDex}~\citep{hoque2025egodexlearningdexterousmanipulation} test datasets (about 7 hours); 
(2) \textbf{Ego4D}~\citep{grauman2022ego4d} test datasets (about 6.5 hours) with VITRA~\citep{li2025vitra} annotations;  
(3) an 10 hours \textbf{in-house collection} (RealSense L515, Figure~\ref{fig:pipeline_evaluation} a); and 
(4) a small \textbf{paired AR-smartphone set} (Figure~\ref{fig:pipeline_evaluation} b). 
Dataset annotation comparisons utilize raw sensor data. 

Evaluation metrics include: 
(1) \textbf{Calibration:} Deviation from offline checkerboard methods~\citep{furgale2013unified}; 
(2) \textbf{Hand Pose:} PA-MPJPE (7-DoF), MPJPE, and AUC; and 
(3) \textbf{Trajectory:} ATE (7-DoF), ATE-S (scale-free 6-DoF), and relative pose errors (RPE).

\textbf{Quantitative Analysis of Calibration.} 
Factory intrinsics (Camera2 API) deviate from offline target-based calibration~\citep{furgale2013unified} by $<1\%$ (mean $0.64\%$, std $0.21\%$). Radial distortion coefficients remain stable at $10^{-3}$ magnitude, validating the reliability of factory parameters for spatial processing without manual recalibration.

\textbf{3D Hand Pose Estimation.} 
Our pipeline achieves high accuracy across benchmarks (Table 1 of Figure~\ref{fig:pipeline_evaluation} f). While larger errors on EgoDex stem primarily from dataset-inherent motion blur and misalignment (Figure~\ref{fig:pipeline_evaluation} d), validation against hardware-tracked AR glasses confirms our model's precision (Figure~\ref{fig:pipeline_evaluation} c). High AUC scores ($>0.90$) further attest to robust keypoint detection across diverse scenarios.

\begin{figure}[h]
    \centering
    \includegraphics[width=0.6\linewidth]{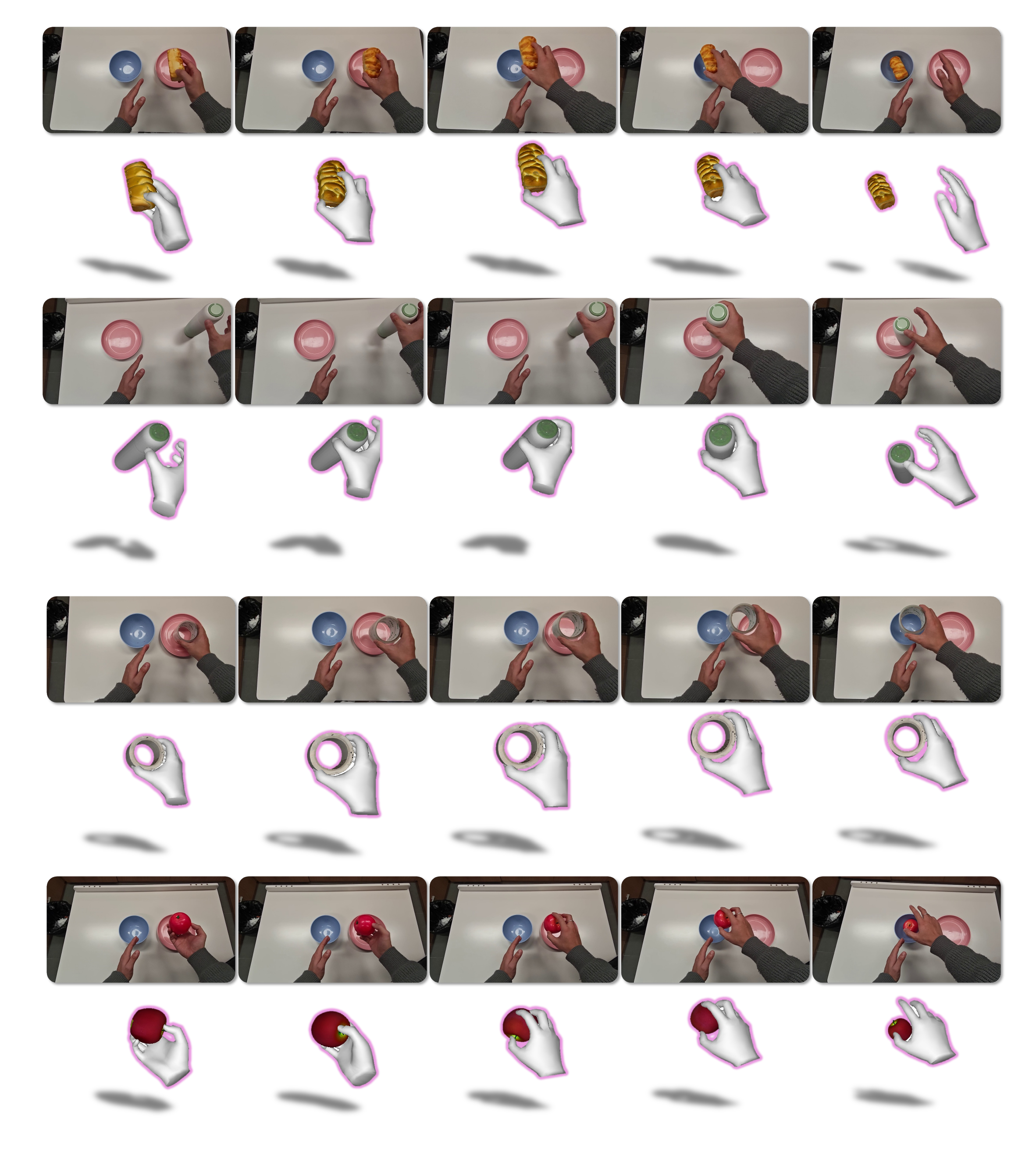}
    \caption{\textbf{Real-to-Sim Reconstruction.}}
    \label{figure:hoi_reconstruction}
\end{figure}

\textbf{Camera Trajectory Estimation.} 
Post 7-DoF alignment, the Absolute Trajectory Error (ATE) remains $<5$~mm across all datasets (Table 2 of Figure~\ref{fig:pipeline_evaluation} f). Although texture-less backgrounds in EgoDex degrade ATE-S ($16.2$~mm), our method maintains centimeter-level accuracy even with monocular RGB input (Figure~\ref{fig:pipeline_evaluation} e), demonstrating robust tracking independent of depth sensors.

\subsection{Real-to-Sim Transferability}

To assess the real-to-sim transferability of \textbf{\textit{AoE}} data, we reconstruct simulation-ready digital twins using the \textbf{AGILE} framework~\citep{shi2026agile}. This process targets the extraction of dynamic hand-object interactions (HOI) from monocular videos, prioritizing interaction fidelity over sparse pose estimation. As illustrated in Figure~\ref{figure:hoi_reconstruction}, our pipeline successfully converts ``in-the-wild'' \textbf{\textit{AoE}} footage into high-fidelity 3D assets. Quantitative evaluation confirms physical validity, achieving a mean penetration depth of $<2~mm$ and robust contact stability—crucial prerequisites for RL-based policy pre-training. These results establish \textbf{\textit{AoE}} as a scalable engine for converting low-cost human demonstrations into high-value simulation assets for dexterous manipulation.

\subsection{Real-World Evaluation on Humanoid Hardware}
To assess the utility of the \textbf{\textit{AoE}} video data for dexterous manipulation, we conduct real-world experiments on a humanoid platform to investigate whether scalable human demonstrations can effectively augment limited robot-specific data and bridge the embodiment gap.

\paragraph{System Configuration.} Experiments utilize a Unitree G1 humanoid with Inspire 5-fingered hands, providing a high-DoF action space. We adopt GR00T N1.5~\citep{bjorck2025gr00t} as the foundation model and employ the FLARE framework~\citep{zheng2025flare} for cross-embodiment fine-tuning. This architecture enables policy co-training using a combination of high-fidelity robot teleoperation data and our \textbf{\textit{AoE}} human video demonstrations.

\begin{figure*}[tp]
\centering
\includegraphics[width=1.0\textwidth]{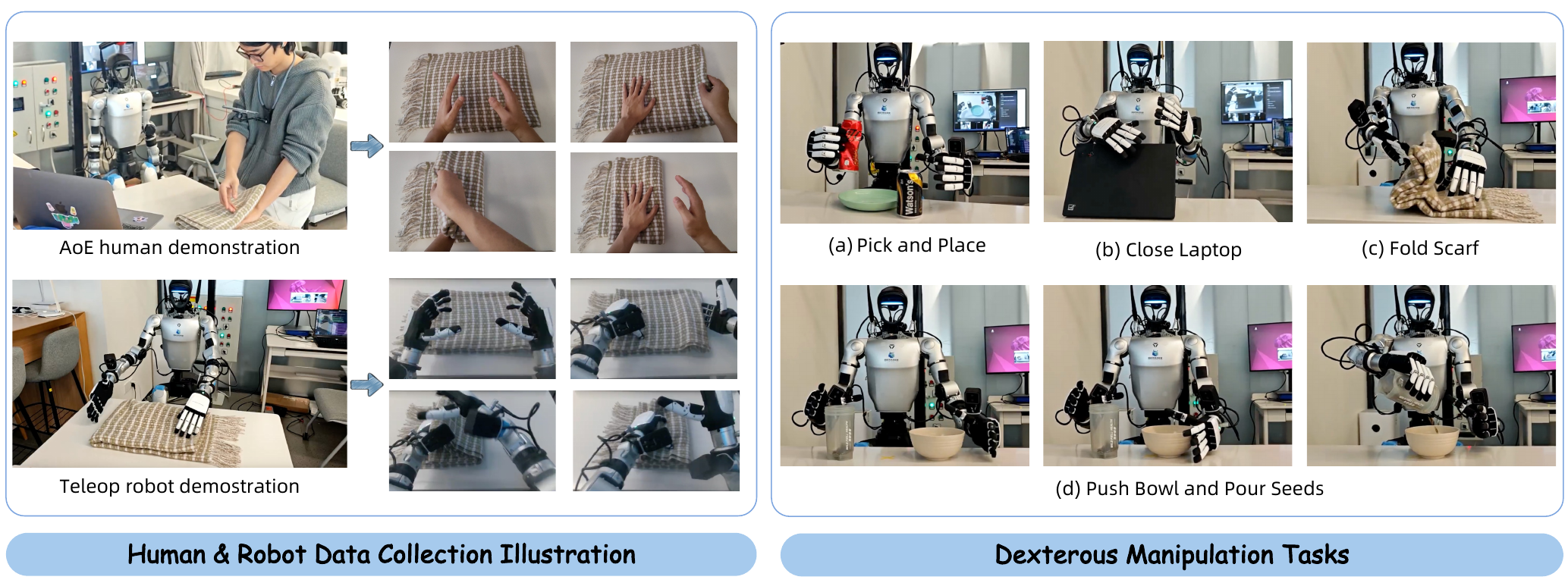}
\caption{
\textbf{Visualization of the real-world experimental setup.} We evaluate the robot's ability to perform diverse dexterous tasks by leveraging \textbf{\textit{AoE}} human demonstrations to augment limited teleoperation data.}
\label{figure:realworld-exp}
\vspace{-15pt} %
\end{figure*}

\paragraph{Task Suite.} We evaluate on four representative tasks (Fig.~\ref{figure:realworld-exp}) encompassing a spectrum of manipulation challenges: (a) \textit{Pick and Place}, requiring precise instruction-following; (b) \textit{Close Laptop}, involving articulated kinematics; (c) \textit{Fold Scarf}, for deformable object handling; and (d) \textit{Push Bowl \& Pour Seeds}, a long-horizon bimanual task requiring fine motor coordination.

\paragraph{Metrics.} Performance is evaluated using Success Rate (SR) and Phase-wise Success Rate (PSR). We systematically vary the ratio of human-to-robot data to quantify sample efficiency gains afforded by \textbf{\textit{AoE}} in low-data regimes. Specifically, we compare a baseline using only 50 robot teleoperation episodes ("50 Teleop") against an augmented setup ("50 Teleop + 200 \textit{AoE}"), where the baseline is supplemented with 200 human video demonstrations from \textit{AoE} for each corresponding task.

\begin{table}[tp]
\centering
\caption{
\textbf{Performance comparison on real-world tasks.} \textbf{SR} and \textbf{PSR} denote Success Rate and Phase-wise Success Rate, respectively. The "+ 200 \textit{AoE}" notation indicates that 200 human video demonstrations of the same task were used to augment the 50 robot teleoperation trials. The results demonstrate that \textbf{\textit{AoE}} data significantly improves performance across various levels of task complexity.}
\label{tab:main_experiment_results}
\resizebox{\textwidth}{!}{
\begin{tabular}{lcccccc}
\toprule
\textbf{Data Recipe} & \textbf{Pick \& Place} & \textbf{Fold Scarf} & \multicolumn{2}{c}{\textbf{Close Laptop}} & \multicolumn{2}{c}{\textbf{Push Bowl \& Pour Seeds}} \\
\cmidrule(lr){2-2} \cmidrule(lr){3-3} \cmidrule(lr){4-5} \cmidrule(lr){6-7}
& SR & SR & SR & PSR & SR & PSR \\
\midrule
50 Teleop & 45.0\% & 10.0\% & 45.0\% & 66.3\% & 0.0\% & 30.0\% \\
50 Teleop + 200 \textbf{\textit{AoE}} & \textbf{75.0\%} & 10.0\% & \textbf{95.0\%} & \textbf{97.5\%} & \textbf{20.0\%} & \textbf{48.0\%} \\
\bottomrule
\end{tabular}
}
\vspace{-15pt} %
\end{table}

\subsubsection{Main Results and Comparative Analysis}
Table~\ref{tab:main_experiment_results} summarizes performance across tasks, indicating that incorporating \textbf{\textit{AoE}} data consistently enhances outcomes, particularly for complex spatial reasoning. Notably, in \textit{Close Laptop}, adding \textbf{\textit{AoE}} data boosts SR from 45.0\% to 95.0\%. In the challenging \textit{Push Bowl \& Pour Seeds} task, where the baseline fails (0\% SR), \textbf{\textit{AoE}} inclusion enables a 20.0\% SR and increases PSR from 30.0\% to 48.0\%. These gains suggest \textbf{\textit{AoE}} provides critical structural priors facilitating long-horizon skill learning. Conversely, performance on \textit{Fold Scarf} remains limited (10.0\% SR); we attribute this bottleneck to hardware latency constraints that hinder the high-frequency reactive control necessary for deformable object manipulation.

\subsubsection{Scaling Behavior and Data Recipes}

\begin{figure}[h]
    \centering
    \includegraphics[width=0.6\linewidth]{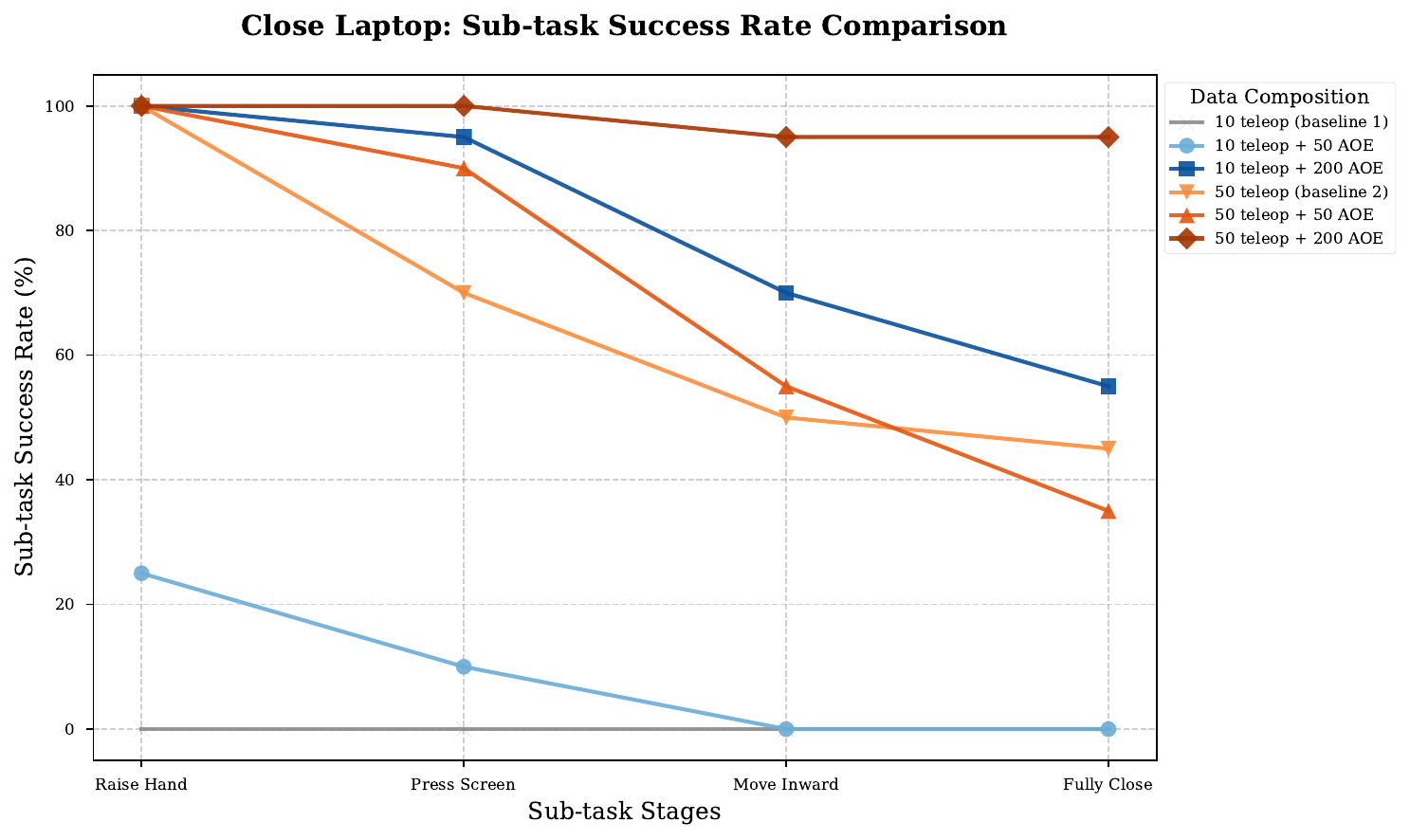}
    \caption{\textbf{Ablation on data recipes for the ``Close Laptop'' task.}}
    \label{figure:ablation_chart}
\end{figure}

To isolate the impact of \textbf{\textit{AoE}} data, we ablate performance on the ``Close Laptop'' task across four kinematic stages, as shown in Figure~\ref{figure:ablation_chart}. First, in sparse data regimes (10 teleop episodes), baselines fail entirely (0\% progress), whereas augmenting with 200 \textbf{\textit{AoE}} episodes bootstraps the policy to 55\% SR, compensating for limited embodiment-specific data. Second, increasing \textbf{\textit{AoE}} volume yields consistent performance gains independent of the robot data baseline. These findings confirm that \textbf{\textit{AoE}} acts as a robust scaling factor, leveraging cross-embodiment demonstrations to significantly enhance policy reliability.

%% file: Sections/5_conclusion.tex
\section{Ethics Statement}

The \textbf{\textit{AoE}} system prioritizes privacy and ethical compliance for always-on egocentric data collection. We adopt a ``privacy-by-design'' architecture where all initial processing (triggering, filtering) occurs on-device; raw data upload requires explicit user authorization. Users retain full control to pause recording, review clips, and withdraw consent at any time. To mitigate re-identification risks, we enforce strict data minimization and de-identification (e.g., blurring faces and screens) prior to upload. Server-side storage employs encryption and strict access controls. The dataset is restricted solely to embodied AI research, explicitly prohibiting surveillance applications. Comprehensive user consent forms and privacy protocols are detailed in Appendix~\ref{appendix:privacy_details}.

\section{Conclusion and Future Work}
In this work, we presented \textbf{\textit{Always-on Egocentric (AoE)}}, a scalable data collection system designed to address the data scarcity bottleneck in embodied foundation models. By leveraging ubiquitous smartphones and human agents, \textbf{\textit{AoE}} circumvents complex hardware dependencies, enabling low-cost, scene-agnostic acquisition of high-quality interaction data. Our cloud-edge collaborative architecture facilitates distributed collection by \textit{anyone, anytime, anywhere}, while automated pipelines transform raw egocentric video into training-ready assets. Extensive evaluations demonstrate that \textbf{\textit{AoE}} data exhibits strong distributional diversity and significantly enhances model generalization in real-world manipulation tasks.

Looking ahead, we plan to release a large-scale, open-source egocentric video dataset built upon the \textbf{\textit{AoE}} system to the research community. Furthermore, we will explore pre-training larger-scale embodied foundation models, investigating how massive, diverse human demonstration data can unlock emergent capabilities in general-purpose robotic agents.

%% file: Sections/6_Appendix.tex
\section{Appendix}

\begin{figure*}[t]
\centering
\includegraphics[width=1.0\textwidth]{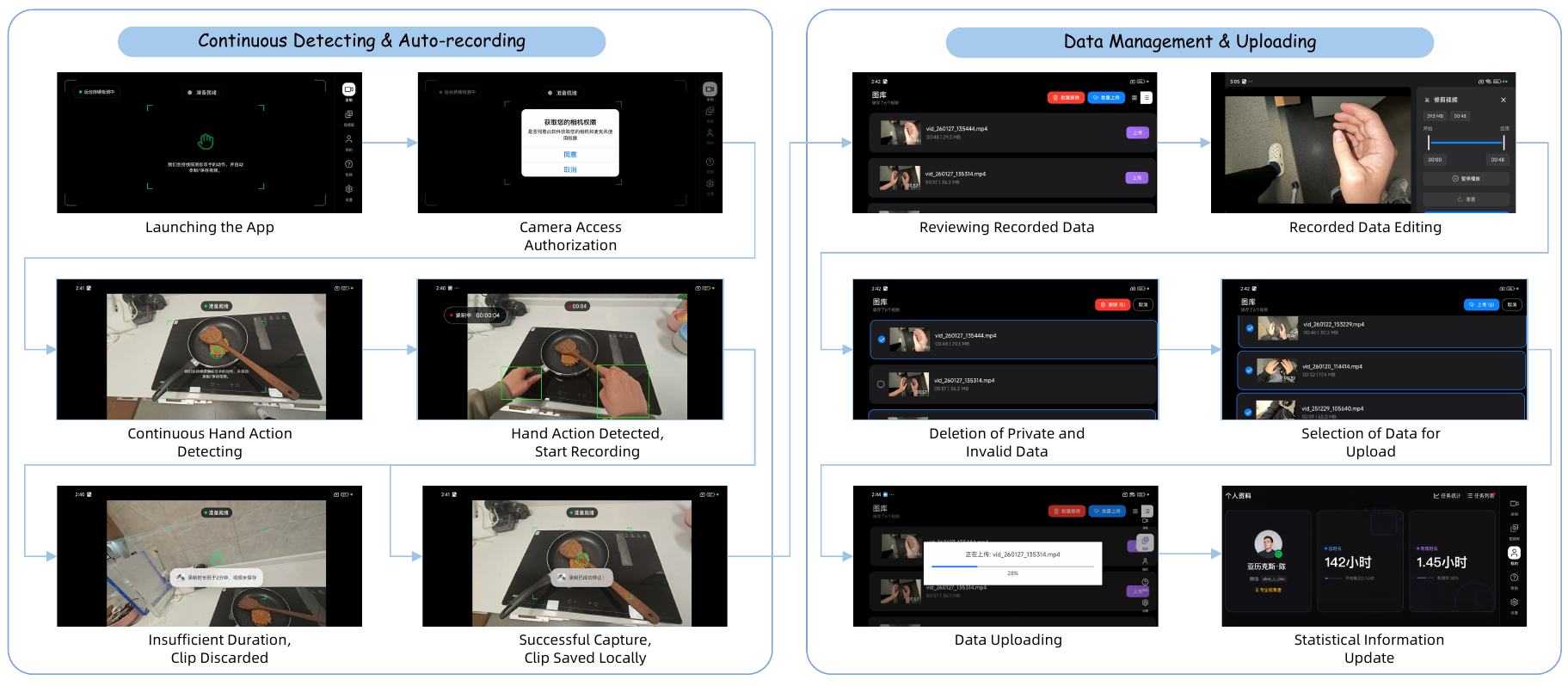}
\caption{
Always-On Collection App Workflow.
}
\vspace{-10pt}
\label{figure:app-workflow}
\end{figure*}

\subsection{Detailed Rating Criteria}
\label{appendix:rating_criteria}

This section details the metrics and evaluation standards used to compare data collection paradigms in Table~\ref{tab:collection_comparison}. We assess each method based on its potential to scale toward the ``foundation model'' era of robotics.

\paragraph{Cost (per user):} This metric represents the total financial investment required to equip a single participant for high-quality data collection. \textbf{Teleoperation} involves the highest barrier ($>\$50\text{k}$) due to industrial arms and specialized master--slave setups. \textbf{UMIs} (\$300--\$800) require GoPro cameras and customized 3D-printed hardware. \textbf{Wearables} ($>\$2\text{k}$) typically involve high-end VR headsets or motion-capture gloves. \textbf{Passive Videos} are considered free as they leverage existing internet resources. \textbf{AOE} ($<\$20$) achieves ultra-low cost by utilizing a user's existing smartphone or action camera paired with a minimal, low-cost mounting solution.

\paragraph{Non-Intrusiveness:} This measures how naturally a user can perform tasks without being hindered by hardware. \textbf{Teleoperation} (\stars{1}) is highly intrusive, often requiring the user to be tethered to a specific station. \textbf{UMIs} (\stars{3}) require the user to hold a specific tool instead of using their hands directly. \textbf{Wearables} (\stars{1}--\stars{2}) often involve heavy headsets or restrictive suits. \textbf{Passive Videos} (\stars{5}) represent natural daily life. \textbf{\textit{AoE}} (\stars{4}) allows for hands-free, natural dexterous manipulation, though the presence of a lightweight mount prevents a perfect score.

\paragraph{Scalability:} This evaluates the feasibility of expanding the data collection pool to thousands of non-expert users globally. Methods requiring expensive hardware (\textbf{Teleoperation}, \textbf{Wearables}) have low scalability (\stars{1}--\stars{2}). \textbf{UMIs} (\stars{3}) are moderately scalable but still require shipping hardware. \textbf{Passive Videos} and \textbf{\textit{AoE}} (\stars{5}) are the most scalable because they rely on ubiquitous consumer electronics (smartphones) or existing web data.

\paragraph{Deployment Ease:} This reflects the technical friction involved in setting up the system. \textbf{Teleoperation} and \textbf{UMIs} (\stars{1}--\stars{2}) require complex calibration, 3D printing, or specific mechanical assembly. \textbf{\textit{AoE}} (\stars{5}) is designed as a ``plug-and-play'' solution, requiring only a simple attachment and an app, comparable to the ease of uploading \textbf{Passive Videos}.

\paragraph{Data Quality:} This assesses the utility of the data for training Vision-Language-Action (VLA) models, particularly for learning fine-grained manipulation. \textbf{Teleoperation} (\stars{5}) provides ``gold-standard'' direct action labels. \textbf{Passive Videos} (\stars{1}) suffer from severe noise, motion blur, and a lack of consistent viewpoints or action labels. \textbf{\textit{AoE}} (\stars{4}) provides high-fidelity egocentric views with consistent hand--object interaction dynamics, providing much cleaner signals for policy learning than raw passive video.

\subsection{Always-On Collection App Workflow}
\label{appendix:app_workflow}

The AOE application workflow, as illustrated in Fig.~\ref{figure:app-workflow}, is designed to balance autonomous data acquisition with user-controlled privacy. The process begins with explicit camera authorization, enabling the app to enter a continuous monitoring state. In this mode, on-device lightweight models scan for hand-object interactions, automatically triggering high-quality recording when relevant actions are detected. To optimize storage and data relevance, the system autonomously discards clips with insufficient duration while locally saving successful captures that meet the quality threshold.

Following the collection phase, the app provides a suite of management tools for data curation. Users can browse the local gallery and utilize a built-in editor to trim irrelevant or sensitive frames from the recordings. To ensure privacy and data integrity, the app requires users to manually select and validate clips before batch-uploading them to the cloud along with synchronized sensor metadata. Finally, the system updates a personal dashboard, providing statistical feedback on total collection hours and effective interaction time to help users monitor their data contribution efficiency.

\subsection{Distributed System Implementation}
\label{appendix:distributed_system}

Large-scale robot learning from human demonstrations requires processing massive volumes of egocentric video data collected from geographically distributed edge devices. To support this objective, we designed a distributed edge-cloud collaborative architecture that addresses three critical system-level challenges:

\textbf{System Design Challenges.} Existing centralized data collection systems face fundamental limitations when scaling to global deployment:
\begin{itemize}
\item \textbf{High-Latency Data Transfer}: Long-distance uploads from geographically dispersed collectors introduce significant network latency (often exceeding 500ms round-trip time) and bandwidth bottlenecks, degrading user experience and limiting upload throughput to impractical levels for high-resolution video streams.
\item \textbf{Inflexible Processing Pipelines}: Hardcoded processing workflows cannot accommodate rapidly evolving egocentric video analysis algorithms. Integrating new algorithms typically requires weeks of system re-engineering, creating a critical bottleneck for algorithmic innovation.
\item \textbf{Limited Scalability}: Centralized architectures lack the ability to dynamically allocate computational resources based on real-time workload, resulting in either resource waste during idle periods or processing delays during peak demands. This limitation severely constrains concurrent data streaming from thousands of devices.
\end{itemize}

To overcome these challenges, our distributed system embodies a decoupled design philosophy across three primary dimensions:
\begin{itemize}
\item \textbf{Geographic Decoupling}: Proximity-based edge ingestion nodes minimize data transfer latency while maintaining centralized data management through asynchronous synchronization.
\item \textbf{Algorithmic Decoupling}: Declarative pipeline orchestration enables rapid integration of new processing algorithms without system-level modifications.
\item \textbf{Resource Decoupling}: Cloud-native elastic scaling dynamically provisions computational resources based on real-time demand, ensuring efficient utilization across varying workloads.
\end{itemize}

The following sections detail our solutions to each challenge: edge-cloud architecture for low-latency data transfer, customizable processing pipeline for algorithmic flexibility, and elastic scaling mechanisms for dynamic resource management.

\subsubsection{Edge-Cloud Collaborative Architecture}

Our system adopts a two-tier architecture that balances local access efficiency with centralized processing power, as illustrated in Figure~\ref{figure:server-pipline}.

\textbf{Distributed Proximity-Based Edge Ingestion.} To minimize data transfer latency, we deploy edge ingestion nodes across multiple geographic regions. The system architecture includes:
\begin{itemize}
\item \textbf{Intelligent Routing}: Collector devices—including smartphones and wearable devices—automatically route to the nearest ingestion point through DNS-based geographic routing and latency-aware selection algorithms.
\item \textbf{Asynchronous Cross-Region Synchronization}: Data collected at each edge node is aggregated to centralized cloud storage through an hourly asynchronous replication mechanism. This design ensures unified data management while maintaining local access efficiency and tolerating temporary network partitions.
\end{itemize}

\textbf{Achieved Benefits.} Compared to traditional centralized architectures, our edge-cloud design delivers measurable improvements:
\begin{itemize}
\item \textbf{Significant Latency Reduction}: Proximity-based edge ingestion reduces average upload latency from 500ms+ to under 100ms, significantly improving user experience and enabling seamless "always-on" data collection.
\item \textbf{High-Concurrency Device Support}: The distributed architecture supports high-concurrency data streaming from thousands of devices simultaneously, enabling truly global-scale data collection.
\end{itemize}

\subsubsection{Customizable Data Processing Pipeline}

To address the challenge of \textbf{Inflexible Processing Pipelines}, we designed a highly flexible pipeline orchestration framework. Egocentric video data processing for robot learning is a rapidly evolving field where optimal algorithms and processing paradigms remain undefined. Existing systems typically hardcode specific processing workflows, forcing researchers to either accept suboptimal methods or invest weeks in system re-engineering for each new algorithm.

\textbf{Custom Operators and Flexible Orchestration.} Our system supports user-defined data processing operators and flexible workflow orchestration through declarative configurations, as illustrated in Figure~\ref{figure:annotation-pipline}. The framework provides three key capabilities:
\begin{itemize}
\item \textbf{Modular Algorithm Integration}: Researchers can rapidly compose different vision algorithm modules (such as hand pose estimation, object detection and tracking, depth reconstruction, action segmentation, etc.) to build end-to-end data processing pipelines according to experimental needs. Each algorithm is encapsulated as an independent operator with standardized input/output interfaces.
\item \textbf{Declarative Workflow Configuration}: Users define custom processing workflows, specifying operator sequences, resource requirements, and data dependencies. This declarative approach eliminates the need for low-level system programming and enables rapid pipeline reconfiguration.
\item \textbf{Hot-Swappable Components}: The system supports runtime replacement of individual operators without pipeline restart, enabling A/B testing of different algorithm versions and seamless algorithm upgrades.
\end{itemize}

\textbf{Achieved Benefits.} This programmability reduces the integration and validation cycle for new algorithms from traditional weeks to days, greatly accelerating algorithm iteration speed.

\subsubsection{Elastic Scaling and Resource Management}

To address the challenge of \textbf{Limited Scalability}, we designed a cloud-native elastic scaling architecture that dynamically provisions computational resources based on real-time demand. Traditional centralized systems allocate fixed resources, leading to either resource waste during idle periods or processing delays during peak demands—a critical limitation when handling concurrent data streams from thousands of globally distributed devices.

\textbf{Dynamic Resource Allocation.} The system employs a cloud-native resource scheduling architecture built on Kubernetes with the following capabilities:
\begin{itemize}
    \item \textbf{Horizontal Pod Autoscaling (HPA)}: The system monitors queue depth and processing latency metrics in real-time, automatically scaling worker nodes based on demand.
    \item \textbf{Intelligent Resource Partitioning}: GPU-intensive operators (e.g., depth estimation, hand reconstruction) are scheduled on GPU nodes, while CPU-bound operators (e.g., data I/O, format conversion) run on cost-effective CPU nodes. This heterogeneous scheduling optimizes both performance and cost.
\end{itemize}

\textbf{Achieved Benefits.} Our elastic scaling architecture delivers:
\begin{itemize}
    \item \textbf{High Resource Utilization}: Dynamic scaling maintains high resource utilization efficiency across varying workloads, avoiding both over-provisioning waste and under-provisioning bottlenecks.
    \item \textbf{Rapid Response Time}: The system responds to workload spikes within minutes, enabling seamless transition from pilot studies to large-scale production without manual capacity planning.
\end{itemize}

\subsection{Experiment Setups}
\label{appendix:experoments}

We employ FLARE \cite{zheng2025flare} to enable robotic policies to learn from human videos without action labels, such as \textbf{\textit{AoE}} data. Unlike teleoperation-based learning paradigms, FLARE \cite{zheng2025flare} is tasked with aligning latent representations of future human actions, ensuring predicted feature embeddings closely match the actual feature vectors of future video frames. This enables the model to learn the physical laws and task logic governing transitions from current to future states by observing vast amounts of human video, even without explicit action labels. In our model design, we transfer this understanding of environmental dynamics to the robot's policy learning.

We fine-tune a pre-trained GR00T N1.5 \cite{bjorck2025gr00t} using the FLARE \cite{zheng2025flare} loss. When training jointly with teleoperation and \textbf{\textit{AoE}} data, we apply supervision via future latent alignment loss and action flow matching loss; when using teleoperation data alone, we apply supervision solely via action flow matching loss. During training, we employed fine-tuning steps that scaled positively with dataset size and task difficulty to prevent overfitting. For instance, for the scarf-folding task, we tested models fine-tuned for 100,000 steps on the largest dataset (50 Teleop + 200 \textbf{\textit{AoE}}). Conversely, for simpler tasks like closing a laptop, we employed only 60,000 fine-tuning steps for the same dataset size.

\subsection{User Consent Forms and Privacy Protocols}
\label{appendix:privacy_details}

\subsubsection*{Overview} This Privacy Policy Summary outlines how \textbf{\textit{AoE}} collects, uses, stores, and protects your personal information when you use our platform services, which include video data collection through first-person perspective hand movement recording.

\subsubsection*{Key Definitions} \begin{itemize}[leftmargin=*] \item \textbf{Personal Information:} Any information recorded electronically or otherwise that relates to an identified or identifiable natural person, excluding anonymized data. \item \textbf{Video Data:} Video files you record and upload through our platform, documenting hand movements from a first-person perspective. \end{itemize}

\subsubsection*{Information We Collect}

\paragraph*{1. Account and Registration Information} \begin{itemize}[leftmargin=*] \item \textbf{Required Information:} Real name, social media account details (with real-name verification) \item \textbf{Purpose:} User identity verification and payment settlement \end{itemize}

\paragraph*{2. Service Usage Information} \begin{itemize}[leftmargin=*] \item \textbf{Video Data:} Hand movement videos you record and upload \item \textbf{Technical Data:} Device information, usage logs, platform interactions \item \textbf{Payment Information:} Identity and payment details for fee settlement \end{itemize}

\paragraph*{3. Digital Certificate Information (if applicable)} \begin{itemize}[leftmargin=*] \item \textbf{Basic Data:} Age, demographic information \item \textbf{Certificate Data:} Academic, driver's license, or vehicle license information \item \textbf{Verification Data:} Identity verification results and supporting documents \end{itemize}

\paragraph*{4. Platform Interaction Data} \begin{itemize}[leftmargin=*] \item \textbf{User Content:} Information you voluntarily share or input \item \textbf{Communication Data:} Customer service interactions, feedback \item \textbf{Technical Logs:} System access records, error reports \end{itemize}

\subsubsection*{How We Use Your Information}

\paragraph*{1. Service Provision and Management} \begin{itemize}[leftmargin=*] \item \textbf{Account Management:} User authentication, account security \item \textbf{Service Delivery:} Video data processing, review, and payment \item \textbf{Technical Support:} Platform maintenance and troubleshooting \end{itemize}

\paragraph*{2. Video Data Processing} \begin{itemize}[leftmargin=*] \item \textbf{Content Review:} Assessment against usability standards \item \textbf{Quality Control:} Technical quality verification \item \textbf{Payment Processing:} Fee calculation and disbursement \end{itemize}

\paragraph*{3. Platform Improvement} \begin{itemize}[leftmargin=*] \item \textbf{User Experience:} Service optimization and personalization \item \textbf{Feature Development:} New functionality research and implementation \item \textbf{Security Enhancement:} Fraud prevention and system protection \end{itemize}

\paragraph*{4. Legal and Compliance} \begin{itemize}[leftmargin=*] \item \textbf{Regulatory Requirements:} Compliance with applicable laws \item \textbf{Dispute Resolution:} Investigation and resolution of user issues \item \textbf{Legal Obligations:} Response to legal requests and proceedings \end{itemize}

\subsubsection*{Information Sharing and Disclosure}

\paragraph*{1. Third-Party Service Providers} \begin{itemize}[leftmargin=*] \item \textbf{Payment Processing:} Financial service partners for fee settlement \item \textbf{Technical Services:} Infrastructure and support providers \item \textbf{Business Partners:} Collaborative service delivery partners \end{itemize}

\paragraph*{2. Legal Requirements} \begin{itemize}[leftmargin=*] \item \textbf{Government Authorities:} When required by law or regulation \item \textbf{Legal Proceedings:} In response to valid legal requests \item \textbf{Protection of Rights:} To protect our legal rights and property \end{itemize}

\paragraph*{3. Business Transfers} \begin{itemize}[leftmargin=*] \item \textbf{Mergers and Acquisitions:} As part of business restructuring \item \textbf{Asset Transfers:} During sale or transfer of business assets \end{itemize}

\subsubsection*{Your Rights and Choices}

\paragraph*{1. Access and Control} \begin{itemize}[leftmargin=*] \item \textbf{Account Settings:} Manage your profile and preferences \item \textbf{Information Access:} Review your personal data \item \textbf{Opt-Out Choices:} Control marketing communications \end{itemize}

\paragraph*{2. Privacy Preferences} \begin{itemize}[leftmargin=*] \item \textbf{Permissions Management:} Control access to device features \item \textbf{Data Sharing:} Manage third-party data sharing \item \textbf{Communication Preferences:} Choose how we contact you \end{itemize}

\paragraph*{3. Account Management} \begin{itemize}[leftmargin=*] \item \textbf{Profile Updates:} Modify your account information \item \textbf{Account Deletion:} Request account termination \item \textbf{Data Portability:} Request your data in accessible format \end{itemize}

\subsubsection*{Data Security and Protection}

\paragraph*{1. Security Measures} \begin{itemize}[leftmargin=*] \item \textbf{Technical Safeguards:} Encryption, access controls, firewalls \item \textbf{Organizational Measures:} Staff training, security policies \item \textbf{Procedural Controls:} Regular security assessments and audits \end{itemize}

\paragraph*{2. Data Retention} \begin{itemize}[leftmargin=*] \item \textbf{Service Data:} Retained for as long as your account is active \item \textbf{Legal Requirements:} Retained to comply with legal obligations \item \textbf{Business Needs:} Retained for legitimate business purposes \end{itemize}

\paragraph*{3. Incident Response} \begin{itemize}[leftmargin=*] \item \textbf{Breach Notification:} Prompt notification of security incidents \item \textbf{Response Procedures:} Established incident management protocols \item \textbf{User Protection:} Measures to protect affected users \end{itemize}

\subsubsection*{Important Privacy Considerations}

\paragraph*{1. Video Data Processing} \begin{itemize}[leftmargin=*] \item \textbf{Content Restrictions:} Avoid including personal information in videos \item \textbf{Rights Transfer:} Complete intellectual property rights transfer upon upload \item \textbf{Usage Rights:} Broad platform rights to use uploaded content \end{itemize}

\paragraph*{2. Cross-Border Data Transfers} \begin{itemize}[leftmargin=*] \item \textbf{Global Operations:} Data may be processed in multiple jurisdictions \item \textbf{International Standards:} Compliance with applicable international standards \item \textbf{User Awareness:} Acknowledge cross-border data processing implications \end{itemize}

\subsubsection*{Platform-Specific Information}

\paragraph*{1. Device Requirements} \begin{itemize}[leftmargin=*] \item \textbf{Collection Device:} Wearable phone holder provided by platform \item \textbf{Personal Device:} Self-provided smartphone with technical specifications \item \textbf{Device Security:} User responsibility for proper use and protection \end{itemize}

\paragraph*{2. Video Standards and Requirements} \begin{itemize}[leftmargin=*] \item \textbf{Content Guidelines:} Production-value hand movements only \item \textbf{Technical Specifications:} Minimum quality and duration requirements \item \textbf{Usage Restrictions:} Prohibited content types and activities \end{itemize}

\paragraph*{3. Fee Settlement and Payments} \begin{itemize}[leftmargin=*] \item \textbf{Eligibility Requirements:} Minimum weekly video duration threshold \item \textbf{Payment Processing:} Third-party payment service providers \item \textbf{Tax Responsibility:} User obligation for tax declaration and payment \end{itemize}

\subsubsection*{Updates and Changes}

\paragraph*{1. Policy Updates} \begin{itemize}[leftmargin=*] \item \textbf{Modification Process:} Periodic review and updates \item \textbf{Notification Methods:} Platform announcements and direct notifications \item \textbf{User Acceptance:} Continued use constitutes acceptance of changes \end{itemize}

\paragraph*{2. Effective Dates} \begin{itemize}[leftmargin=*] \item \textbf{Current Policy:} Applies from specified effective date \item \textbf{Future Changes:} Take effect according to specified timelines \item \textbf{Historical Versions:} Previous versions available upon request \end{itemize}

\subsubsection*{Contact and Support}

\paragraph*{1. Privacy Inquiries} \begin{itemize}[leftmargin=*] \item \textbf{Contact Methods:} Official hotline and support channels \item \textbf{Response Times:} Reasonable timeframe for inquiries \item \textbf{Complaint Resolution:} Process for addressing privacy concerns \end{itemize}

\paragraph*{2. Technical Support} \begin{itemize}[leftmargin=*] \item \textbf{Device Assistance:} Help with collection device usage \item \textbf{Platform Issues:} Technical problem resolution \item \textbf{Account Support:} Assistance with account management \end{itemize}

\paragraph*{3. Legal and Compliance} \begin{itemize}[leftmargin=*] \item \textbf{Regulatory Questions:} Information about legal obligations \item \textbf{Data Protection:} Details about privacy rights and protections \item \textbf{Contractual Terms:} Clarification of user agreement provisions \end{itemize}

\vspace{2em} \noindent\textbf{Important Notice:} This summary provides an overview of key privacy considerations but does not replace the full User Agreement. Please read the complete agreement for detailed terms and conditions. By using our services, you acknowledge and agree to our data processing practices as described in our full privacy policy and user agreement.

\vspace{1em} \noindent\textit{Effective Date:} As specified in the complete User Agreement\ \textit{Last Updated:} \today

\vspace{1em} \noindent\textbf{Your continued use of \textbf{\textit{AoE}} services constitutes acceptance of our privacy practices and terms of service.}

%% file: main.bib
@article{chi2024universal,
  title={Universal manipulation interface: In-the-wild robot teaching without in-the-wild robots},
  author={Chi, Cheng and Xu, Zhenjia and Pan, Chuer and Cousineau, Eric and Burchfiel, Benjamin and Feng, Siyuan and Tedrake, Russ and Song, Shuran},
  journal={arXiv preprint arXiv:2402.10329},
  year={2024}
}

@inproceedings{zhaxizhuoma2025fastumi,
  title={FastUMI: A scalable and hardware-independent universal manipulation interface with dataset},
  author={Zhaxizhuoma, Zhaxizhuom and Liu, Kehui and Guan, Chuyue and Jia, Zhongjie and Wu, Ziniu and Liu, Xin and Wang, Tianyu and Liang, Shuai and CHEN, Pengan and Zhang, Pingrui and others},
  booktitle={Conference on Robot Learning},
  pages={3069--3093},
  year={2025},
  organization={PMLR}
}

@article{zeng2025activeumi,
  title={ActiveUMI: Robotic Manipulation with Active Perception from Robot-Free Human Demonstrations},
  author={Zeng, Qiyuan and Li, Chengmeng and John, Jude St and Zhou, Zhongyi and Wen, Junjie and Feng, Guorui and Zhu, Yichen and Xu, Yi},
  journal={arXiv preprint arXiv:2510.01607},
  year={2025}
}

@inproceedings{xu2025exumi,
  title     = {exUMI: Extensible Robot Teaching System with Action-aware Task-agnostic Tactile Representation},
  author    = {Xu, Yue and Wei, Litao and An, Pengyu and Zhang, Qingyu and Li, Yong-Lu},
  booktitle = {Conference on Robot Learning (CoRL)},
  year      = {2025}
}

@article{zhong2025humanoidexo,
  title={HumanoidExo: Scalable Whole-Body Humanoid Manipulation via Wearable Exoskeleton},
  author={Zhong, Rui and Sun, Yizhe and Wen, Junjie and Li, Jinming and Cheng, Chuang and Dai, Wei and Zeng, Zhiwen and Lu, Huimin and Zhu, Yichen and Xu, Yi},
  journal={arXiv preprint arXiv:2510.03022},
  year={2025}
}

@inproceedings{lindata,
  title={Data Scaling Laws in Imitation Learning for Robotic Manipulation},
  author={Lin, Fanqi and Hu, Yingdong and Sheng, Pingyue and Wen, Chuan and You, Jiacheng and Gao, Yang},
  booktitle={The Thirteenth International Conference on Learning Representations}
}

@article{generalist2025gen0,
author = {GeneralistAI},
title = {GEN-0: Embodied Foundation Models That Scale with Physical Interaction},
journal = {Generalist AI Blog},
year = {2025},
note = {https://generalistai.com/blog/preview-uqlxvb-bb.html},
}

@inproceedings{chen2025towards,
  title={Towards Effective Utilization of Mixed-Quality Demonstrations in Robotic Manipulation via Segment-Level Selection and Optimization},
  author={Chen, Jingjing and Fang, Hongjie and Fang, Hao-Shu and Lu, Cewu},
  booktitle={2025 IEEE International Conference on Robotics and Automation (ICRA)},
  pages={16884--16891},
  year={2025},
  organization={IEEE}
}

@inproceedings{luo2025human,
  title={Human-agent joint learning for efficient robot manipulation skill acquisition},
  author={Luo, Shengcheng and Peng, Quanquan and Lv, Jun and Hong, Kaiwen and Driggs--Campbell, Katherine Rose and Lu, Cewu and Li, Yong--Lu},
  booktitle={2025 IEEE International Conference on Robotics and Automation (ICRA)},
  pages={1370--1377},
  year={2025},
  organization={IEEE}
}

@inproceedings{xu2025dexumi,
  title={DexUMI: Using Human Hand as the Universal Manipulation Interface for Dexterous Manipulation},
  author={Xu, Mengda and Zhang, Han and Hou, Yifan and Xu, Zhenjia and Fan, Linxi and Veloso, Manuela and Song, Shuran},
  booktitle={3rd RSS Workshop on Dexterous Manipulation: Learning and Control with Diverse Data}
}

@inproceedings{wang2024dexcap,
  title={DexCap: Scalable and Portable Mocap Data Collection System for Dexterous Manipulation},
  author={Wang, Chen and Shi, Haochen and Wang, Weizhuo and Zhang, Ruohan and Fei-Fei, Li and Liu, Karen},
  booktitle={RSS 2024 Workshop: Data Generation for Robotics}
}

@inproceedings{fang2024airexo,
  title={Airexo: Low-cost exoskeletons for learning whole-arm manipulation in the wild},
  author={Fang, Hongjie and Fang, Hao-Shu and Wang, Yiming and Ren, Jieji and Chen, Jingjing and Zhang, Ruo and Wang, Weiming and Lu, Cewu},
  booktitle={2024 IEEE International Conference on Robotics and Automation (ICRA)},
  pages={15031--15038},
  year={2024},
  organization={IEEE}
}

@inproceedings{grauman2022ego4d,
  title={Ego4d: Around the world in 3,000 hours of egocentric video},
  author={Grauman, Kristen and Westbury, Andrew and Byrne, Eugene and Chavis, Zachary and Furnari, Antonino and Girdhar, Rohit and Hamburger, Jackson and Jiang, Hao and Liu, Miao and Liu, Xingyu and others},
  booktitle={Proceedings of the IEEE/CVF conference on computer vision and pattern recognition},
  pages={18995--19012},
  year={2022}
}

@inproceedings{damen2018scaling,
  title={Scaling egocentric vision: The epic-kitchens dataset},
  author={Damen, Dima and Doughty, Hazel and Farinella, Giovanni Maria and Fidler, Sanja and Furnari, Antonino and Kazakos, Evangelos and Moltisanti, Davide and Munro, Jonathan and Perrett, Toby and Price, Will and others},
  booktitle={Proceedings of the European conference on computer vision (ECCV)},
  pages={720--736},
  year={2018}
}

@inproceedings{mavip,
  title={VIP: Towards Universal Visual Reward and Representation via Value-Implicit Pre-Training},
  author={Ma, Yecheng Jason and Sodhani, Shagun and Jayaraman, Dinesh and Bastani, Osbert and Kumar, Vikash and Zhang, Amy},
  booktitle={The Eleventh International Conference on Learning Representations}
}

@inproceedings{nair2023r3m,
  title={R3M: A Universal Visual Representation for Robot Manipulation},
  author={Nair, Suraj and Rajeswaran, Aravind and Kumar, Vikash and Finn, Chelsea and Gupta, Abhinav},
  booktitle={Conference on Robot Learning},
  pages={892--909},
  year={2023},
  organization={PMLR}
}

@inproceedings{radosavovic2023real,
  title={Real-world robot learning with masked visual pre-training},
  author={Radosavovic, Ilija and Xiao, Tete and James, Stephen and Abbeel, Pieter and Malik, Jitendra and Darrell, Trevor},
  booktitle={Conference on Robot Learning},
  pages={416--426},
  year={2023},
  organization={PMLR}
}

@inproceedings{bharadhwaj2024track2act,
  title={Track2act: Predicting point tracks from internet videos enables generalizable robot manipulation},
  author={Bharadhwaj, Homanga and Mottaghi, Roozbeh and Gupta, Abhinav and Tulsiani, Shubham},
  booktitle={European Conference on Computer Vision},
  pages={306--324},
  year={2024},
  organization={Springer}
}

@article{wen2023any,
  title={Any-point trajectory modeling for policy learning},
  author={Wen, Chuan and Lin, Xingyu and So, John and Chen, Kai and Dou, Qi and Gao, Yang and Abbeel, Pieter},
  journal={arXiv preprint arXiv:2401.00025},
  year={2023}
}

@inproceedings{wangmimicplay,
  title={MimicPlay: Long-Horizon Imitation Learning by Watching Human Play},
  author={Wang, Chen and Fan, Linxi and Sun, Jiankai and Zhang, Ruohan and Fei-Fei, Li and Xu, Danfei and Zhu, Yuke and Anandkumar, Anima},
  booktitle={7th Annual Conference on Robot Learning}
}

@inproceedings{qin2022dexmv,
  title={Dexmv: Imitation learning for dexterous manipulation from human videos},
  author={Qin, Yuzhe and Wu, Yueh-Hua and Liu, Shaowei and Jiang, Hanwen and Yang, Ruihan and Fu, Yang and Wang, Xiaolong},
  booktitle={European Conference on Computer Vision},
  pages={570--587},
  year={2022},
  organization={Springer}
}

@article{chen2022dextransfer,
  title={Dextransfer: Real world multi-fingered dexterous grasping with minimal human demonstrations},
  author={Chen, Zoey Qiuyu and Van Wyk, Karl and Chao, Yu-Wei and Yang, Wei and Mousavian, Arsalan and Gupta, Abhishek and Fox, Dieter},
  journal={arXiv preprint arXiv:2209.14284},
  year={2022}
}

@inproceedings{chen2025vividex,
  title={Vividex: Learning vision-based dexterous manipulation from human videos},
  author={Chen, Zerui and Chen, Shizhe and Arlaud, Etienne and Laptev, Ivan and Schmid, Cordelia},
  booktitle={2025 IEEE International Conference on Robotics and Automation (ICRA)},
  pages={3336--3343},
  year={2025},
  organization={IEEE}
}

@inproceedings{yelatent,
  title={Latent Action Pretraining from Videos},
  author={Ye, Seonghyeon and Jang, Joel and Jeon, Byeongguk and Joo, Se June and Yang, Jianwei and Peng, Baolin and Mandlekar, Ajay and Tan, Reuben and Chao, Yu-Wei and Lin, Bill Yuchen and others},
  booktitle={The Thirteenth International Conference on Learning Representations}
}

@article{bu2025univla,
  title={Univla: Learning to act anywhere with task-centric latent actions},
  author={Bu, Qingwen and Yang, Yanting and Cai, Jisong and Gao, Shenyuan and Ren, Guanghui and Yao, Maoqing and Luo, Ping and Li, Hongyang},
  journal={arXiv preprint arXiv:2505.06111},
  year={2025}
}

@article{bjorck2025gr00t,
  title={Gr00t n1: An open foundation model for generalist humanoid robots},
  author={Bjorck, Johan and Casta{\~n}eda, Fernando and Cherniadev, Nikita and Da, Xingye and Ding, Runyu and Fan, Linxi and Fang, Yu and Fox, Dieter and Hu, Fengyuan and Huang, Spencer and others},
  journal={arXiv preprint arXiv:2503.14734},
  year={2025}
}

@inproceedings{mandikal2022dexvip,
  title={Dexvip: Learning dexterous grasping with human hand pose priors from video},
  author={Mandikal, Priyanka and Grauman, Kristen},
  booktitle={Conference on Robot Learning},
  pages={651--661},
  year={2022},
  organization={PMLR}
}

@article{yang2025egovla,
  title={Egovla: Learning vision-language-action models from egocentric human videos},
  author={Yang, Ruihan and Yu, Qinxi and Wu, Yecheng and Yan, Rui and Li, Borui and Cheng, An-Chieh and Zou, Xueyan and Fang, Yunhao and Cheng, Xuxin and Qiu, Ri-Zhao and others},
  journal={arXiv preprint arXiv:2507.12440},
  year={2025}
}

@article{luo2025being,
  title={Being-h0: vision-language-action pretraining from large-scale human videos},
  author={Luo, Hao and Feng, Yicheng and Zhang, Wanpeng and Zheng, Sipeng and Wang, Ye and Yuan, Haoqi and Liu, Jiazheng and Xu, Chaoyi and Jin, Qin and Lu, Zongqing},
  journal={arXiv preprint arXiv:2507.15597},
  year={2025}
}

@article{bi2025h,
  title={H-rdt: Human manipulation enhanced bimanual robotic manipulation},
  author={Bi, Hongzhe and Wu, Lingxuan and Lin, Tianwei and Tan, Hengkai and Su, Zhizhong and Su, Hang and Zhu, Jun},
  journal={arXiv preprint arXiv:2507.23523},
  year={2025}
}

@article{cai2025n,
  title={In-N-On: Scaling Egocentric Manipulation with in-the-wild and on-task Data},
  author={Cai, Xiongyi and Qiu, Ri-Zhao and Chen, Geng and Wei, Lai and Liu, Isabella and Huang, Tianshu and Cheng, Xuxin and Wang, Xiaolong},
  journal={arXiv preprint arXiv:2511.15704},
  year={2025}
}

@article{kim2025fine,
          title={Fine-Tuning Vision-Language-Action Models: Optimizing Speed and Success},
          author={Kim, Moo Jin and Finn, Chelsea and Liang, Percy},
          journal={arXiv preprint arXiv:2502.19645},
          year={2025}
}

@article{lingbot-depth2026,
  title={Masked Depth Modeling for Spatial Perception},
  author={Tan, Bin and Sun, Changjiang and Qin, Xiage and Adai, Hanat and Fu, Zelin and Zhou, Tianxiang and Zhang, Han and Xu, Yinghao and Zhu, Xing and Shen, Yujun and Xue, Nan},
  journal={arXiv preprint arXiv:2601.17895},
  year={2026}
}

@article{Qwen2.5-VL,
  title={Qwen2.5-VL Technical Report},
  author={Bai, Shuai and Chen, Keqin and Liu, Xuejing and Wang, Jialin and Ge, Wenbin and Song, Sibo and Dang, Kai and Wang, Peng and Wang, Shijie and Tang, Jun and Zhong, Humen and Zhu, Yuanzhi and Yang, Mingkun and Li, Zhaohai and Wan, Jianqiang and Wang, Pengfei and Ding, Wei and Fu, Zheren and Xu, Yiheng and Ye, Jiabo and Zhang, Xi and Xie, Tianbao and Cheng, Zesen and Zhang, Hang and Yang, Zhibo and Xu, Haiyang and Lin, Junyang},
  journal={arXiv preprint arXiv:2502.13923},
  year={2025}
}

@article{li2025vitra,
  title={Scalable Vision-Language-Action Model Pretraining for Robotic Manipulation with Real-Life Human Activity Videos},
  author={Qixiu Li and Yu Deng and Yaobo Liang and Lin Luo and Lei Zhou and Chengtang Yao and Lingqi Zeng and Zhiyuan Feng and Huizhi Liang and Sicheng Xu and Yizhong Zhang and Xi Chen and Hao Chen and Lily Sun and Dong Chen and Jiaolong Yang and Baining Guo},
  journal={arXiv preprint arXiv:2510.21571},
  year={2025}
}

@manual{android2024camera2overview,
  title = {{Camera2 API} Overview - {Android} Developers},
  author = {{Google LLC}},
  year = {2024},
  url = {https://developer.android.com/training/camera2},
  note = {Accessed: 2025-02-06}
}

@article{furgale2013unified,
  title={Unified temporal and spatial calibration for multi-sensor systems},
  author={Furgale, Paul and Barfoot, Timothy D and Sibley, Gabe},
  booktitle={2013 IEEE/RSJ International Conference on Intelligent Robots and Systems},
  year={2013}
}

@article{wang2025moge2,
      title={MoGe-2: Accurate Monocular Geometry with Metric Scale and Sharp Details}, 
      author={Ruicheng Wang and Sicheng Xu and Yue Dong and Yu Deng and Jianfeng Xiang and Zelong Lv and Guangzhong Sun and Xin Tong and Jiaolong Yang},
      year={2025}
}

@article{li2024megasam,
  title={{MegaSaM}: Accurate, Fast, and Robust Structure and Motion from Casual Dynamic Videos},
  author={Li, Zhengqi and Tucker, Richard and Cole, Forrester and Wang, Qianqian and Jin, Linyi and Ye, Vickie and Kanazawa, Angjoo and Holynski, Aleksander and Snavely, Noah},
  journal={arXiv preprint arXiv:2412.04463},
  year={2024},
}

@article{zhang2025hawor,
      title={HaWoR: World-Space Hand Motion Reconstruction from Egocentric Videos},
      author={Zhang, Jinglei and Deng, Jiankang and Ma, Chao and Potamias, Rolandos Alexandros},
      journal={arXiv preprint arXiv:2501.02973},
      year={2025}
}

@article{Romero_2017,
   title={Embodied hands: modeling and capturing hands and bodies together},
   volume={36},
   ISSN={1557-7368},
   url={http://dx.doi.org/10.1145/3130800.3130883},
   DOI={10.1145/3130800.3130883},
   number={6},
   journal={ACM Transactions on Graphics},
   publisher={Association for Computing Machinery (ACM)},
   author={Romero, Javier and Tzionas, Dimitrios and Black, Michael J.},
   year={2017},
   month=nov, pages={1–17} }

@article{lepert2025masquerade,
    title={Masquerade: Learning from In-the-wild Human Videos using Data-Editing},
    author={Lepert, Marion and Fang, Jiaying and Bohg, Jeannette},
    journal={arXiv preprint arXiv:2508.09976},
    year={2025}
}

@misc{hoque2025egodexlearningdexterousmanipulation,
      title={EgoDex: Learning Dexterous Manipulation from Large-Scale Egocentric Video}, 
      author={Ryan Hoque and Peide Huang and David J. Yoon and Mouli Sivapurapu and Jian Zhang},
      year={2025},
      eprint={2505.11709},
      archivePrefix={arXiv},
      primaryClass={cs.CV},
      url={https://arxiv.org/abs/2505.11709}, 
}

@article{zheng2025flare,
  title={FLARE: Robot learning with implicit world modeling},
  author={Zheng, Ruijie and Wang, Jing and Reed, Scott and Bjorck, Johan and Fang, Yu and Hu, Fengyuan and Jang, Joel and Kundalia, Kaushil and Lin, Zongyu and Magne, Loic and others},
  journal={arXiv preprint arXiv:2505.15659},
  year={2025}
}

@inproceedings{ramanishka2018CVPR,    
   author={Vasili Ramanishka and Yi-Ting Chen and Teruhisa Misu and Kate Saenko},    
   title={Toward Driving Scene Understanding: A Dataset for Learning Driver Behavior and Causal Reasoning},    
   booktitle={Conference on Computer Vision and Pattern Recognition (CVPR)},    
   year={2018}
}

@article{mao2021one,
  title={One million scenes for autonomous driving: Once dataset},
  author={Mao, Jiageng and Niu, Minzhe and Jiang, Chenhan and Liang, Hanxue and Chen, Jingheng and Liang, Xiaodan and Li, Yamin and Ye, Chaoqiang and Zhang, Wei and Li, Zhenguo and others},
  journal={arXiv preprint arXiv:2106.11037},
  year={2021}
}

@article{qiu2025humanpolicy,
    title={Humanoid Policy \~{} Human Policy},
    author={Ri-Zhao Qiu and Shiqi Yang and Xuxin Cheng and Chaitanya Chawla and Jialong Li and Tairan He and Ge Yan and David J. Yoon and Ryan Hoque and Lars Paulsen and Ge Yang and Jian Zhang and Sha Yi and Guanya Shi and Xiaolong Wang},
    journal={arXiv preprint arXiv:2503.13441},
    year={2025}
  }

@article{shi2026agile,
  title={AGILE: Hand-Object Interaction Reconstruction from Video via Agentic Generation},
  author={Shi, Jin-Chuan and Ye, Binhong and Liu, Tao and He, Junzhe and Xu, Yangjinhui and Liu, Xiaoyang and Li, Zeju and Chen, Hao and Shen, Chunhua},
  journal={arXiv preprint arXiv:2602.04672},
  year={2026}
}
